%% file: iclr2021_conference.tex
\documentclass{article} % For LaTeX2e
\usepackage{iclr2021_conference,times}

\input{math_commands.tex}

\usepackage{hyperref}
\usepackage{url}
\usepackage{adjustbox}
\usepackage{wrapfig}

\usepackage{algorithm}
\usepackage{algorithmic}
\floatstyle{ruled}
\newfloat{algorithm}{tbp}{loa}
\algsetup{linenodelimiter=}

\newcommand{\din}{\mathcal D}
\newcommand{\dout}{\mathcal D^{\prime}}
\newcommand{\task}{\mathcal {T}}

\newcommand{\RMAMLa}{\text{R-MAML$_{\mathrm{both}}$}}
\newcommand{\RMAMLb}{\text{R-MAML$_{\mathrm{out}}$}}
\newcommand{\modelR}{\mathbf w_{\mathrm{r}}}
\newcommand{\modelC}{\mathbf w_{\mathrm{c}}}
\newcommand{\RMAMLbAnil}{\text{R-MAML$_{\mathrm{out}}$-ANIL}}
\newcommand{\RMAMLbFgsm}{\text{R-MAML$_{\mathrm{out}}$-FGSM}}

\newcommand{\RMAMLbTr}{\text{R-MAML$_{\mathrm{out}}$-TRADES}}
\newcommand{\RMAMLbCL}{\text{R-MAML$_{\mathrm{out}}$-CL}}

\DeclareMathOperator*{\minimize}{\text{minimize}}
\DeclareMathOperator*{\maximize}{\text{maximize}}

\DeclareMathOperator*{\st}{\text{subject to}}
\DeclareMathAlphabet\mathbfcal{OMS}{cmsy}{b}{n}

\newcommand{\SL}[1]{\textcolor{red}{SL: #1}}

\newcommand{\mycomment}[1]{}

\newcommand{\tcb}{\textcolor{blue}}

\title{On Fast Adversarial Robustness Adaptation in Model-Agnostic Meta-Learning
}

\author{Ren Wang$^{1,4}$ \thanks{Corresponding: Ren Wang (wangren348117609@gmail.com, renwang@umich.edu). 
}
\quad ~Kaidi Xu$^{2}$~~~~Sijia Liu$^{3,5}$ \thanks{Work is done at the MIT-IBM Watson AI Lab}~~~~Pin-Yu Chen$^{3}$~~~~Tsui-Wei Weng$^{3}$~~~~Chuang Gan$^{3}$\\
\textbf{Meng Wang}$^{1}$\\
    $^1$Rensselaer Polytechnic Institute, USA\\
  $^2$Northeastern University, USA\\
    $^3$MIT-IBM Watson AI Lab, IBM Research, USA\\
    $^4$University of Michigan, USA\\
    $^5$Michigan State University, USA
}

\iclrfinalcopy % Uncomment for camera-ready version, but NOT for submission.
\begin{document}

\maketitle

\begin{abstract}
Model-agnostic meta-learning (MAML) has emerged as one of the most successful meta-learning techniques in few-shot learning. It enables us to learn a  \textit{meta-initialization} of model parameters (that we call \textit{meta-model}) to rapidly adapt to new tasks using a small amount of labeled training data.  %(namely, )
% of great generalization-ability in few-shot learning. That is,  new tasks can be learned by fine-tuning the meta-initialization over a small amount of labeled data. In spite of generalizationability 
Despite the generalization power of the meta-model, it remains elusive that how \textit{adversarial robustness} can be maintained by MAML in few-shot learning. In addition to generalization, robustness is also desired for a meta-model to defend adversarial examples (attacks). 
%\Lily{This sentence feels like generalization is not important. We can say "In addition to generalization, robustness is also desired ..."}
%namely,  examples with imperceptible perturbations can deceive these models and lead to erroneous prediction.
Toward promoting adversarial robustness in MAML, we first study \textit{when} a robustness-promoting regularization should be incorporated, given the fact that MAML adopts a bi-level (fine-tuning vs. meta-update) learning procedure. We show that robustifying the meta-update stage is sufficient to make robustness adapted to the task-specific fine-tuning stage even if the latter uses a standard training protocol. We also make additional justification on the acquired robustness adaptation   by peering into the interpretability of neurons' activation maps. Furthermore, we investigate \textit{how} robust regularization can \textit{efficiently} be designed in MAML. We propose a general but easily-optimized robustness-regularized meta-learning framework, which allows the use of unlabeled data augmentation, fast adversarial attack generation, and computationally-light fine-tuning. In particular, we for the first time show that the auxiliary contrastive learning task can enhance the adversarial robustness of MAML. Finally, extensive experiments are conducted to demonstrate the effectiveness of our proposed methods in robust few-shot learning. Codes are available at \url{https://github.com/wangren09/MetaAdv}.
\end{abstract}

\section{Introduction}
\label{intro}
%\SL{[I will find a slot to revise  tomorrow afternoon.]}

Meta-learning, which can offer fast generalization adaptation to unseen tasks \citep{thrun2012learning,novak1984learning}, has  widely been studied  from 
%gained increasing interest and attention in   literature, ranging from 
model- and metric-based methods \citep{santoro2016meta,munkhdalai2017meta,koch2015siamese,snell2017prototypical}
to   optimization-based methods 
 \citep{ravi2016optimization,finn2017model,nichol2018first}.
% is a category of methods that can rapidly adapt to new tasks that have never been encountered \citep{thrun2012learning,novak1984learning}. The common approaches of meta-learning can be divided into three main categories: Model-based \citep{santoro2016meta,munkhdalai2017meta}, Metric-based \citep{koch2015siamese,snell2017prototypical}, and Optimization-based \citep{finn2017model,nichol2018first}. 
In particular, 
model-agnostic meta-learning (MAML) \citep{finn2017model} is one of the most intriguing bi-level  optimization-based meta-learning methods designed for fast-adapted few-shot learning. That is,
% Specifically,
% MAML contains a two-stage learning process using gradient descent. 
the learnt meta-model can rapidly be  generalized to 
%MAML can rapidly generalize to
unforeseen tasks with only a small amount of data. It has  successfully been applied to   use cases
%is useful in scenarios requiring few-shot learning 
such as object detection \citep{wang2020tracking}, medical image analysis \citep{maicas2018training}, and language modeling \citep{huang2018natural}. 

In addition to generalization-ability,
recent works \citep{yin2018adversarial,goldblum2019adversarially,xu2020yet}  investigated MAML from   another fundamental  perspective, \textit{adversarial robustness}, given by the capabilities of a model defending against adversarially perturbed inputs (known as adversarial examples/attacks) \citep{goodfellow2014explaining, xu2019structured}.
The challenge of lacking robustness of deep learning (DL) models  has gained increasing interest and attention. And there exists a proactive arm race between  adversarial attack and defense; see overview in \citep{carlini2019evaluating,hao2020adversarial}.

%\citep{raghu2019rapid,Wong2020Fast}

There have existed many defensive methods in the context of {standard model training}, e.g., \citep{madry2017towards,zhang2019theoretically,Wong2020Fast,carmon2019unlabeled,stanforth2019labels,xu2019topology}, however, few work studied \textit{robust MAML} except 
\citep{yin2018adversarial,goldblum2019adversarially} to the best of our knowledge. 
And tackling such a problem is more challenging than
%In contrast to 
robustifying the standard model training, since MAML contains a bi-leveled  learning procedure in 
which the meta-update step (outer loop)
optimizes a task-agnostic initialization of model parameters while the fine-tuning step (inner loop) learns a task-specific model instantization updated from the common initialization. Thus, it remains elusive \textit{when} (namely, at which learning stage) and \textit{how} 
robust regularization should be promoted to strike a graceful balance between generalization/robustness and computation efficiency. Note that neither the standard MAML \citep{finn2017model}
nor the standard robust training \citep{madry2017towards,zhang2019theoretically} is as easy as normal training. Besides the algorithmic design in robust MAML, it is also important to  draw in-depth explanation and analysis on \textit{why} adversarial robustness can efficiently be gained in MAML. 
In this work, we aim to re-visit 
the problem of adversarial robustness in MAML  \citep{yin2018adversarial,goldblum2019adversarially} and make affirmative answers to the  above  questions on \textit{when}, \textit{how} and \textit{why}. 

%the problem of robust MAML poses 

\paragraph{Contributions}\label{para:contribution}
Compared to the existing works \citep{yin2018adversarial,goldblum2019adversarially}, we make the following contributions:

$\bullet$ Given the fact that MAML is formed as a bi-level learning procedure, we show and explain why regularizing    adversarial robustness at the meta-update level is sufficient to  
  offer fast and effective robustness adaptation on few-shot test tasks.

$\bullet$ Given the fact that either  MAML or robust training alone is computationally intensive, we propose a general but efficient robustness-regularized meta-learning framework, which allows the use of unlabeled data augmentation, fast (one-step) adversarial example generation during meta-updating, 
%{(requiring only one additional gradient descent and one additional backward step compared to MAML)}
and partial model training during fine-tuning {(only fine-tuning the classifier's head)}. 
%\Lily{Maybe can add what is our computation overhead here vs pure MAML.}

$\bullet$ We for the first time show that the use of unlabeled data augmentation, particularly  introducing
an   auxiliary contrastive learning task, can provide additional benefits on
 adversarial robustness of MAML
 in the low data regime, $2\%$ robust accuracy improvement and $9\%$ clean accuracy improvement over the state-of-the-art robust MAML method (named as \textit{adversarial querying}) in \citep{goldblum2019adversarially}. 

\paragraph{Related work}
% \SL{In what follows, we review relevant literature from {robust training}, {efficient MAML},
% and {robust MAML}.
% } 

%\textit{Robust training:}
To train a standard model (instead of a meta-model), the most effective robust training methods include 
adversarial training \citep{madry2017towards}, TRADES that places a theoretically-grounded trade-off between accuracy and robustness \citep{zhang2019theoretically},
and   their many variants such as fast adversarial training methods  \citep{shafahi2019adversarial,zhang2019you,Wong2020Fast,andriushchenko2020understanding},  semi-supervised robust training \citep{carmon2019unlabeled,stanforth2019labels},
adversarial transfer learning %\citep{chen2020adversarial,Shafahi2020Adversarially,chan2020thinks}, 
and certifiably robust training \citep{wong2017provable,dvijotham2018training}.
Moreover, recent works
\citep{hendrycks2019using,chen2020adversarial,Shafahi2020Adversarially,chan2020thinks,utrera2020adversarially,salman2020adversarially} studied the transferability of robustness in  
%to a new task domain has also been studied 
in the context of
transfer learning and representation learning. 
However, the aforementioned standard robust training methods are not directly applicable to MAML in few-shot learning considering MAML's bi-leveled optimization nature.

\mycomment{
\cite{hendrycks2019using} conducts adversarial training on random initialized models and pre-trained models, and shows that pre-training helps improve robustness. In contrast,  \cite{Shafahi2020Adversarially} studies the cases of fine-tuning a robust model using clean validation data. They show that the generalization can be improved by properly selecting the strategy while the robustness can remain the same. It is also shown in \citep{utrera2020adversarially} that the adversarially-trained model transfers better than the standard model. \cite{chan2020thinks} forces the input gradients of a student model to be aligned with the robust teacher model, and finds that the operation improves the robustness. The adversarial transfer learning studied in the above works cannot be generalized to the MAML framework. This is because MAML has a two-stage learning process and requires a few-shot fine-tuning, which are different from the conventional adversarial transfer learning. 
}

A few recent works   studied the problem of adversarial training in the context of MAML   \citep{goldblum2019adversarially,yin2018adversarial}. \cite{yin2018adversarial} considered the robust training in both fine-tuning  and meta-update steps, which is  unavoidably  computationally expensive and difficult in optimization.
The most relevant work to ours is \citep{goldblum2019adversarially}, which proposed 
adversarial querying (AQ) by integrating  
 adversarial training  with MAML.  Similar to ours, 
 AQ attempted to robustify  meta-update only to gain sufficient robustness. However, it lacks explanation for the rationale behind that.  We will show that AQ can also be regarded as a special case of our proposed robustness-promoting MAML framework. Most important, we make a more in-depth study with novelties summarized in
\textbf{Contributions}.

Another line of  research relevant to ours is   efficient MAML, e.g., \citep{raghu2019rapid,song2019maml,su2019does}, where   the goal is to improve the computation efficiency and/or the generalization of MAML.
In \citep{song2019maml}, gradient-free optimization was leveraged to alleviate the need of second-order derivative information during meta-update. In \citep{raghu2019rapid}, MAML was simplified by removing the fine-tuning step over the representation block of a meta-model. It was shown that such a simplification is surprisingly effective without losing generalization-ability.  In \citep{su2019does}, a self-supervised representation learning task  was augmented to the meta-updating objective and resulted in a meta-model with improved   generalization. Although useful insights were gained from MAML in the aforementioned works, none of them took adversarial robustness into account.

\section{Preliminaries and Problem Statement}
In this section, we first review  model-agnostic meta learning (MAML) \citep{finn2017model}   and {adv}ersarial {t}raining \citep{madry2017towards}, respectively. We then motivate the setup of robustness-promoting MAML and demonstrate its  challenges in design 
%due to the existence of a high freedom \Lily{Not sure what "high freedom" means here} 
when integrating MAML with robust regularization. 

\paragraph{MAML}
MAML attempts to learn an initialization of model parameters (namely, a meta-model) so that a   new few-shot task  can quickly and easily be tackled by fine-tuning   this meta-model  over a small amount of labeled data. 
The characteristic signature of MAML is its \textit{bi-level} learning procedure, where the {fine-tuning} stage forms a task-specific \textit{inner loop} while the meta-model is updated at the \textit{outer loop} by minimizing the   validation error of  fine-tuned models over cumulative tasks. 
Formally, 
 consider $N$ few-shot learning tasks $\{ \task_i \}_{i=1}^N$, each of which has a fine-tuning data set $\din_i$ and a validation set $\dout_i$, where $\din_i$ is used in the \textit{fine-tuning} stage and $\dout_i$ is used in the  \textit{meta-update} stage. Here the superscript $(\prime)$ is preserved to indicate   operations/parameters at the meta-upate stage.  
% Let
%  $\ell_i(\mathbf w; \mathcal{D})$
% denote a task-specific loss evaluated on the data set $\mathcal D$ using an ML model (e.g.,  neural network) parameterized by $\mathbf w$. 
MAML is then formulated as the following bi-level optimization problem \citep{finn2017model}:
%\SL{[Will find a better formulation]}
\begin{align}\label{eq: prob_MAML}
   % \hspace*{-0.05in}
    \begin{array}{ll}
\displaystyle \minimize_{\mathbf w}         & \frac{1}{N} \sum_{i=1}^N \ell_i^{\prime}( 
%\mathbf w_i^{(K)}
\mathbf w_i^{\prime}
; \dout_i )  \\
\st & \mathbf w_i^{\prime} = \argmin_{\mathbf w_i} 
\ell_i (\mathbf w_i; \din_i, \mathbf w), ~\forall i \in [N]
    %   \st   & \mathbf w_i^{(k)} =  \mathbf w_i^{(k-1)} - \alpha  \nabla_{\mathbf w_i}  \ell_i( \mathbf w_i^{(k-1)} ; \din_i ), \text{ and }  \mathbf w_i^{(0)} = \mathbf w,  ~ \forall k \in [K], \forall i \in [N], 
    \end{array}
   % \hspace*{-0.15in}
\end{align}
%%% use different function f, g?
where $\mathbf w$ denotes the meta-model to be designed, $\mathbf w_i^{\prime}$ is the $\task_i$-specific fine-tuned model,
$ \ell_i^{\prime}( 
%\mathbf w_i^{(K)}
\mathbf w_i^{\prime}
; \dout_i )$ represents the   validation error using the fine-tuned model, 
$\ell_i(\mathbf w_i; \din_i, \mathbf w)$ denotes the 
 training error when fine-tuning the task-specific model parameters  $\mathbf w_i$ using the task-agnostic initialization  $\mathbf w$,
%fine-tuned over $\din_i$ from the initialization $\mathbf w$
%via $K$-step gradient descent (GD), $\alpha > 0$ is a given step size, 
and for ease of notation, $[K]$ represents the integer set $\{1,2,\ldots,K \}$. In \eqref{eq: prob_MAML}, the objective function and  the constraint correspond to the meta-update stage and  fine-tuning stage, respectively. The bi-level optimization problem is challenging because each constraint calls an
inner optimization oracle, which is typically   instantiated into a $K$-step gradient descent (GD) based solver: 
\[
\mathbf w_i^{(k)} =  \mathbf w_i^{(k-1)} - \alpha  \nabla_{\mathbf w_i}  \ell_i( \mathbf w_i^{(k-1)} ; \din_i , \mathbf w), ~ k \in [K], \text{ with }  \mathbf w_i^{(0)} = \mathbf w.
\]
We note that even with the above simplified fine-tuning step, 
updating the meta-model $\mathbf w$  still requires the second-order derivatives of the objective function  of \eqref{eq: prob_MAML} with respect to  (w.r.t.) $\mathbf w$.
\mycomment{$
\mathbf w_i^{\prime} = \mathbf w_i^{(K)}$ with
$ 
 \mathbf w_i^{(k)} =  \mathbf w_i^{(k-1)} - \alpha  \nabla_{\mathbf w_i}  \ell_i( \mathbf w_i^{(k-1)} ; \din_i, \mathbf w )
$ and  $\mathbf w_i^{(0)} = \mathbf w$ for  $k \in [K]$.
}
% Here the fine-tuned model $\mathbf w_i^{(k)}$ can  be interpreted as a solution to the optimization problem 
% $\min_{\mathbf w_i} \ell_i(\mathbf w_i; \din_i)$ with initialization $\mathbf w_i = \mathbf w$. And 
% Accordingly,
% updating the meta-model $\mathbf w$  at least requires the second-order derivatives of the objective function  of \eqref{eq: prob_MAML} with respect to  (w.r.t.) $\mathbf w$.

%the \underline{i}nner loop of MAML to fine-tune   the  initial optimizee variable $\btheta$ of our interest, and $\dout_i$ is used in the \underline{o}uter loop of MAML to test the fine-tuned model.

\paragraph{Adversarial training}
The min-max optimization based
adversarial training (AT) is known  as one of the most powerful defense methods to obtain a robust model against adversarial attacks \citep{madry2017towards}. %It  has also   been used as a backbone strategy that leads to   many state-of-the-art defenses. 
We summarize AT and its variants through the following robustness-regularized optimization problem: 
\begin{align}\label{eq: prob_robust}
    \begin{array}{ll}
\displaystyle\minimize_{\mathbf w}    &
\lambda  \mathbb E_{(\mathbf x, y) \in \mathcal D} 
\left [ \ell(\mathbf w; \mathbf x, y) \right ] +
\underbrace{\mathbb E_{(\mathbf x, y) \in \mathcal D} [ \displaystyle \maximize_{\| \boldsymbol{\delta} \|_\infty \leq \epsilon} 
 g( \mathbf w; \mathbf x+ \boldsymbol \delta,  y ) ]}_\text{$\mathcal R(\mathbf w; \mathcal D)$},
    \end{array}
\end{align}
where $\ell(\mathbf w; \mathbf x, y)$ denotes the prediction loss evaluated at the point $\mathbf x$ with label $y$, $\lambda \geq 0$ is a regularization parameter,  
$\boldsymbol \delta$ denotes the input perturbation variable within  the $\ell_\infty$-norm ball of radius $\epsilon$,
  $g$ represents the robust loss evaluated at the model $\mathbf w$ at the perturbed example $\mathbf x+ \boldsymbol \delta$ given the true label $y$, and for ease of notation, let $\mathcal R(\mathbf w; \mathcal D)$ denote the robust regularization function for model $\mathbf w$ under the data set $\mathcal D$.
In the rest of the   paper, we consider two specifications of $\mathcal R$: (a) \textit{AT regularization}  \citep{madry2017towards}, where
we set $g  = \ell $   and $\lambda = 0$; (b) \textit{TRADES regularization}  \citep{zhang2019theoretically}, where  we define $ g  $   as  the cross-entropy between the distribution of  prediction probabilities 
%\PY{(logits or softmax outputs? I don't think KL is well-defined on logits)} 
  at the perturbed example $(\mathbf x + \boldsymbol \delta )$ and that   at the original sample $\mathbf x$. 
 %\textbf{Robust MAML is Nontrivial}

%  $\mathbf{z}(\mathbf x; \boldsymbol \theta)$ represents  the  probability distribution   over    class labels predicted by  the  model $\boldsymbol \theta$, and $\mathrm{KL}$ denotes the Kullback–Leibler divergence (or cross-entropy)
%     %\LH{I assume that KL does decent job and there is no need to change to symmetric measures as JS?} 
%     function between the prediction at the perturbed sample $(\mathbf x + \boldsymbol \delta)$ and that at the original sample $\mathbf x$. 
    
% \section{Few-Short Robust Learning Models and Efficient Training Strategies}
% \label{prob_form}

\paragraph{Robustness-promoting MAML}
Integrating  MAML with AT is  a natural solution to enhance     adversarial robustness of a   meta-model in few-shot learning. 
However, this seemingly simple scheme is in fact far from trivial, and there exist three critical roadblocks as elaborated below. 

\textit{First}, it remains elusive at which stage (fine-tuning or meta-update) robustness can most effectively be  gained for MAML. 
Based on \eqref{eq: prob_MAML} and \eqref{eq: prob_robust}, we can cast this problem as a unified optimization problem that  
  augments the MAML  loss   with the robust regularization    under two degrees of freedom characterized by  two hyper-parameters $\gamma_{\mathrm{out}} \geq 0 $ and $\gamma_{\mathrm{in}}  \geq 0$:
\begin{align}\label{eq: prob_MAML_robust}
   % \hspace*{-0.05in}
    \begin{array}{ll}
\displaystyle \minimize_{\mathbf w}         & \frac{1}{N} \sum_{i=1}^N [ \ell_i^{\prime}( \mathbf w_i^{\prime}; \dout_i ) + \gamma_{\mathrm{out}}  \mathcal R_i(\mathbf w_i^{\prime}; \dout_i) ] \\
      \st   &   \mathbf w_i^{\prime} = \argmin_{\mathbf w_i}  [
\ell_i (\mathbf w_i; \din_i, \mathbf w) + \gamma_{\mathrm{in}} \mathcal R_i(\mathbf w_i; \din_i) ], ~\forall i \in [N].
    \end{array}
   % \hspace*{-0.15in}
\end{align}
Here $\mathcal R_i$ denotes the task-specific robustness regularizer, and the choice of $(\gamma_{\mathrm{in}}, \gamma_{\mathrm{out}})$ determines the specific scenario of robustness-promoting MAML.
Clearly,
% there exist many scenarios to specify  problem \eqref{eq: prob_MAML_robust}, determined by the choices of $\gamma_{\mathrm{out}}$ and  $\gamma_{\mathrm{in}}$.  
% The  
the direct application is to set $\gamma_{\mathrm{in}} > 0$ and  $\gamma_{\mathrm{out}} > 0$, that is, both fine-tuning and meta-update steps would be carried out using robust training, which calls additional loops to generate adversarial examples.  Thus, this would make computation most intensive.
%\Lily{I think we can elaborate a bit why including both is computationally-intensive (because have to solve a max problem), and how much expensive it is}. 
Spurred by that, we ask: \textit{Is it possible to achieve a robust meta-model 
%with transferable robustness to fine-tuning tasks 
by 
incorporating robust regularization into only either meta-update or fine-tuning step (corresponding to $\gamma_{\mathrm{ in}} = 0$ or $\gamma_{\mathrm{out}} = 0$)?} %\Lily{should be $\gamma$}
 
\textit{Second},  both MAML in \eqref{eq: prob_MAML} and AT in \eqref{eq: prob_robust} are challenging bi-level optimization problems which need to call inner optimization routines for fine-tuning and attack generation, respectively. 
Thus, we ask whether or not the computationally-light alternatives of inner solvers, e.g., partial fine-tuning \citep{raghu2019rapid} and fast attack generation \citep{Wong2020Fast}, can promise adversarial robustness in few-shot learning. %robustness-aware few-shot meta-learning. 

\textit{Third}, it has been shown that  adversarial robustness can   benefit from semi-supervised learning by leveraging (unlabeled) data augmentation
\citep{carmon2019unlabeled, stanforth2019labels}.
Spurred by that, we further ask: \textit{Is it possible to generalize robustness-promoting MAML to the setup of semi-supervised learning for improved accuracy-robustness tradeoff?}

%     In \eqref{eq:maml-orig}, the fairness regularization is conducted at both the inner and the outer loops.
%     However, it is unknown how fairness is gained across the meta-training and   meta-validation steps. If fair representations 
%      have already been learned from   the updated meta-initialization $\boldsymbol{\theta}$,  
% then the fairness regularization    could be removed from the inner loop. This leads to the simplified setting with standard finetuning

%\section{Adversarial Training Strategies in Meta-Training and Meta-Testing}

\section{When to Incorporate Robust Regularization in MAML?
}
\label{sec: when}

\begin{wrapfigure}{r}{60mm}
\vspace*{-0.2in}
\centerline{
\includegraphics[width=.475\textwidth,height=!]{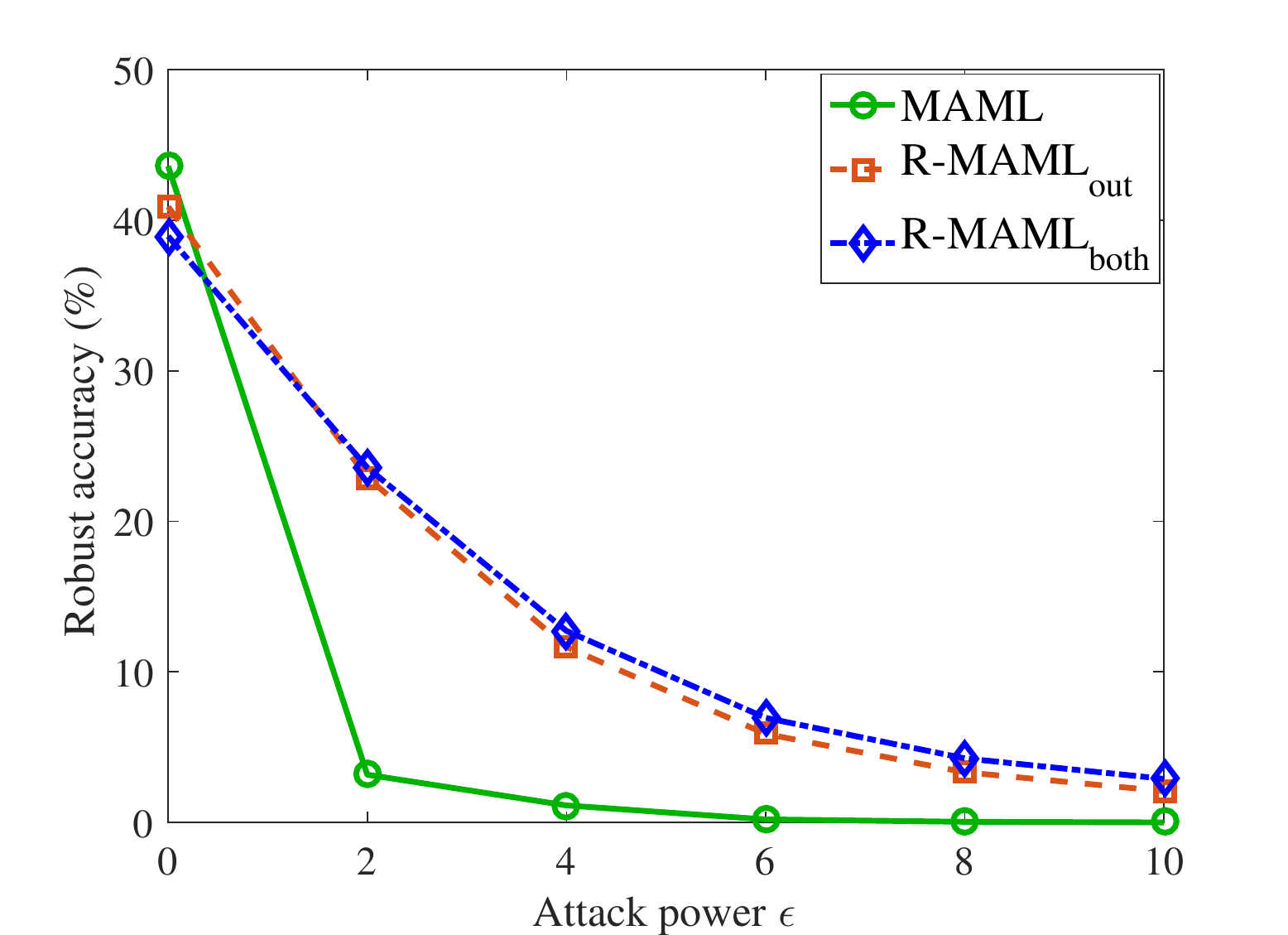}
}
\vspace*{-0.13in}
\caption{\small{
RA of meta-models trained by standard MAML, {\RMAMLa} and {\RMAMLb} versus  PGD attacks of different perturbation sizes during meta-testing. Results show that robustness regularized meta-update with standard fine-tuning (namely, {\RMAMLb}) has 
already been effective in promotion of robustness.
% applied to meta-test tasks 
% RA of different models against various PGD attacks of different perturbation sizes during testing. The meta-model acquired by  {\RMAMLb} yields \textit{nearly   the same robust accuracy} (RA) as  {\RMAMLa}.
}}
% \caption{\small{Comparison of different models with different meta-training strategies in 1-shot 5-way scenario. Robust accuracy decreases when $\epsilon$ increases (10 attack steps). MAML (Adv-MU) represents the robust regularization in the meta-update stage. MAML (Adv-IUMU) represents incorporating the robust regularization in both the inner-gradient update (fine-tuning) and meta-update.}}
  \label{comp_in_out}
  %\vspace*{-0.2in}
\end{wrapfigure}

In this section, we evaluate at which stage adversarial robustness  can be gained during meta-training. We will provide insights and step-by-step investigations to show when to incorporate robust training in MAML and why it works.
Based on \eqref{eq: prob_MAML_robust}, we focus on {two robustness-promoting meta-training protocols}. \textbf{(a)} \RMAMLa, where 
{r}obustness regularization applied to \textit{both}  fine-tuning and meta-update steps with $\gamma_{\mathrm{in}}, \gamma_{\mathrm{out}} > 0$; \textbf{(b)} \RMAMLb, where robust regularization applied to  \textit{meta-update only}, i.e.,   $\gamma_{\mathrm{in}} = 0$ and $\gamma_{\mathrm{out}} > 0$. 
Compared to  {\RMAMLa},     {\RMAMLb} is more user-friendly since it allows the use of
   \textit{standard} fine-tuning    over the learnt robust meta-model when tackling unseen few-shot test tasks (known as meta-testing).
   In what follows, we will show that even if {\RMAMLb} does not use robust regularization in fine-tuning, it is sufficient to warrant the transferability of meta-model's robustness to downstream fine-tuning tasks.

    %It is also worth mentioning that the   protocol (b) was   studied  

\paragraph{All you need is robust meta-update during meta-training}
To study this claim, we solve problem \eqref{eq: prob_MAML_robust} 
%[This equation is too far away from this paper. Might want to pull that up?]
using {\RMAMLa} and {\RMAMLb} respectively   in the   $5$-way $1$-shot  learning setup, where $1$  data sample at each of $5$ randomly selected  MiniImagenet classes   \citep{ravi2016optimization} constructs a learning task. Throughout this section, we specify $\mathcal R_i$ in \eqref{eq: prob_MAML_robust} as the   AT regularization, which   calls a $10$-step projected gradient descent (PGD) attack generation method with $\epsilon = 2/255$ in its inner maximization subroutine given by \eqref{eq: prob_robust}.
We refer readers to Section\,\ref{exp_note} for more implementation details.

We  find that the meta-model acquired by  {\RMAMLb} yields \textit{nearly   the same robust   accuracy} (RA) as  {\RMAMLa} against  various PGD attacks generated at the testing phase using different perturbation sizes    $\epsilon = \{ 0,2,\ldots, 10\}/255$ as shown in Figure\,\ref{comp_in_out}. 
Unless specified otherwise, we evaluate the performance of the
 meta-learning schemes over $2400$ random unseen $5$-way $1$-shot test tasks. 
 We also note that RA under $\epsilon = 0$ becomes the standard accuracy (SA) evaluated using benign (unperturbed) test examples. 
 It is clear from Figure\,\ref{comp_in_out} that
 both {\RMAMLb} and {\RMAMLa}  can yield significantly better RA than MAML with  slightly worse SA. 
It is also expected that RA decreases as the attack power $\epsilon$ increases.  

\begin{wrapfigure}{r}{50mm}
\vspace*{-5mm}
  \centerline{
  \begin{adjustbox}{max width=0.37\textwidth }
    \begin{tabular}{@{\hskip 0.00in}c  @{\hskip 0.00in} c @{\hskip 0.00in}   }
 \begin{tabular}{@{}c@{}}  
\vspace*{0.01in}\\
\rotatebox{90}{\parbox{6.7em}{\centering \normalsize \textbf{Seed Images}}}
% \vspace*{-0.05in}
 \\
\rotatebox{90}{\parbox{6.7em}{\centering \normalsize \textbf{\begin{tabular}[c]{@{}c@{}}IAMs\\ (MAML)  \end{tabular}  }}}
 \\
\rotatebox{90}{\parbox{6.7em}{\centering \normalsize \textbf{\begin{tabular}[c]{@{}c@{}}IAMs\\ ({\RMAMLa})  \end{tabular}}}}
\\
\rotatebox{90}{\parbox{6.7em}{\centering \normalsize \textbf{\begin{tabular}[c]{@{}c@{}}IAMs\\ ({\RMAMLb})  \end{tabular} }}}
\\
\end{tabular} 
&
 \begin{tabular}{@{\hskip 0.01in}c@{\hskip 0.01in}c@{\hskip 0.01in}c@{\hskip 0.01in}c@{\hskip 0.01in}}
\begin{tabular}{@{\hskip 0.01in}c@{\hskip 0.00in}}
%\\
 \parbox[c]{7em}{\includegraphics[width=6.7em]{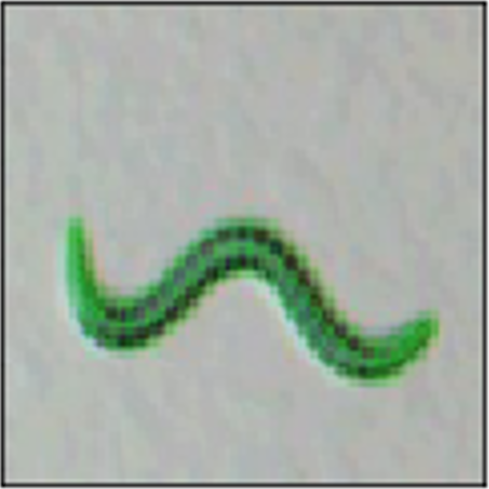}} 
 \\
 \parbox[c]{7em}{\includegraphics[width=6.7em]{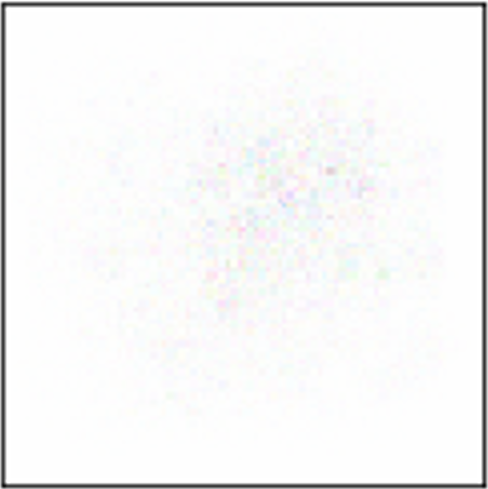}}
  \\
  \parbox[c]{7em}{\includegraphics[width=6.7em]{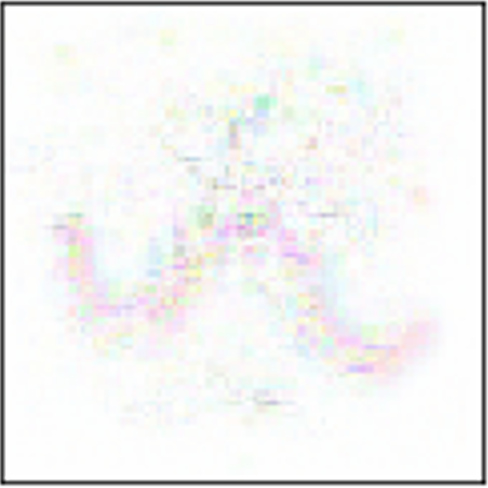}}
  \\
 \parbox[c]{7em}{\includegraphics[width=6.7em]{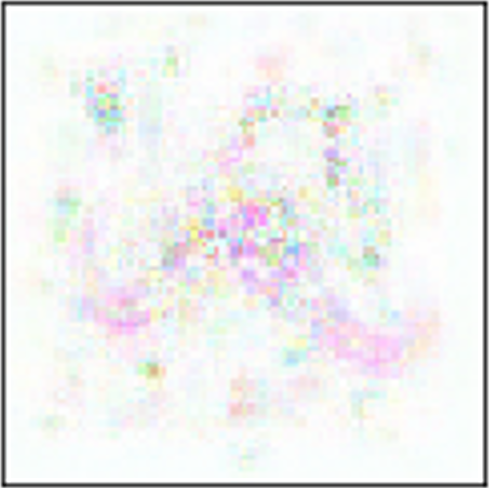}}
 \\
\end{tabular}
&
 \begin{tabular}{@{\hskip 0.01in}c@{\hskip 0.01in}}
%\\
 \parbox[c]{7em}{\includegraphics[width=6.7em]{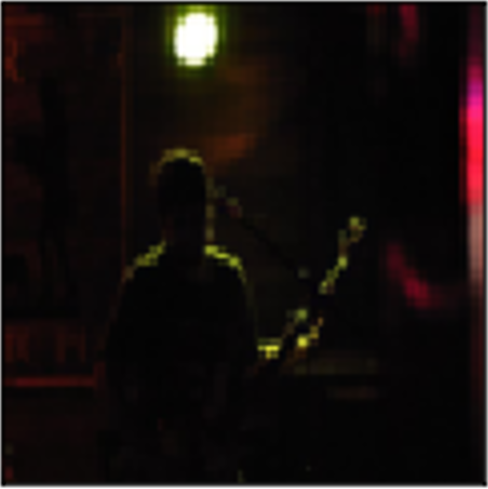}} 
 \\
 \parbox[c]{7em}{\includegraphics[width=6.7em]{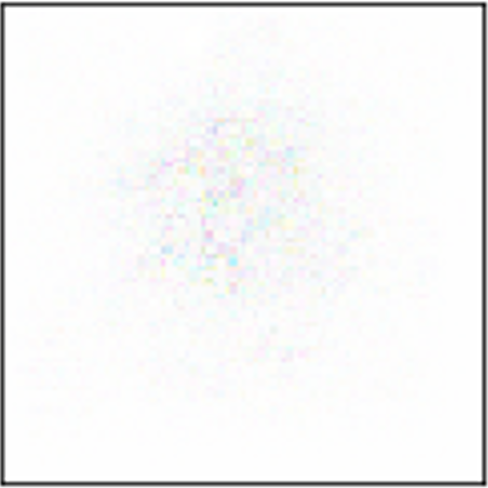}}
  \\
  \parbox[c]{7em}{\includegraphics[width=6.7em]{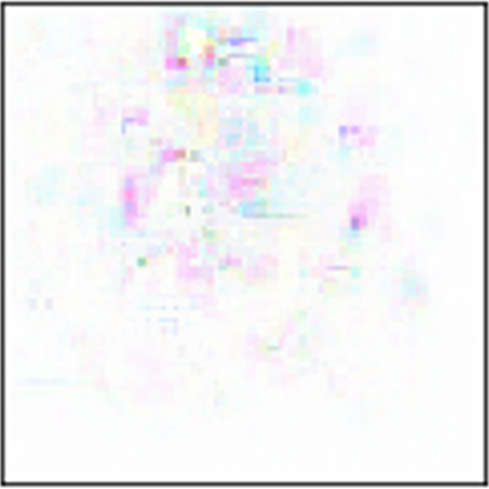}}
  \\
 \parbox[c]{7em}{\includegraphics[width=6.7em]{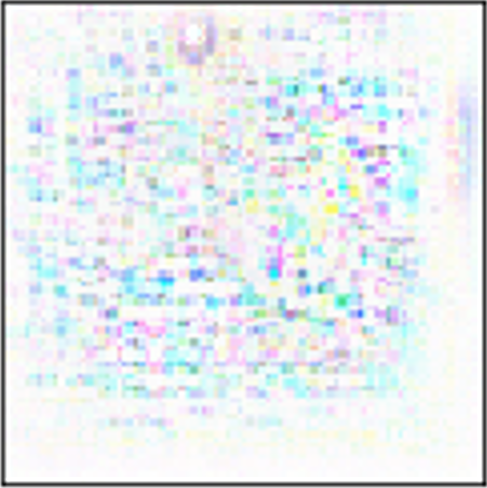}}
 \\
\end{tabular}
 &
 \begin{tabular}{@{\hskip 0.00in}c@{\hskip 0.01in}}
 \parbox[c]{7em}{\includegraphics[width=6.7em]{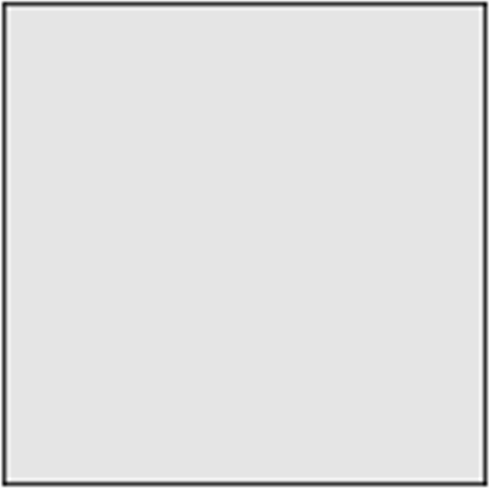}} 
 \\
 \parbox[c]{7em}{\includegraphics[width=6.7em]{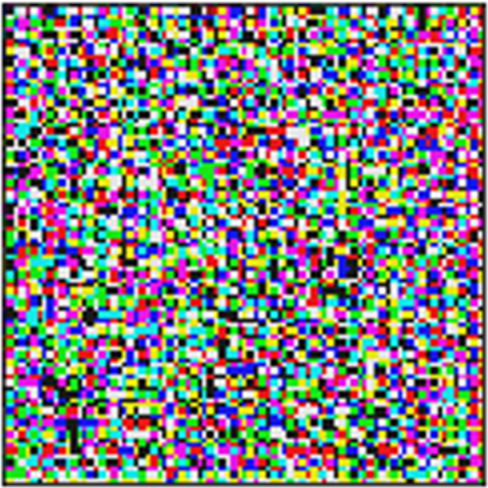}} 
  \\
  \parbox[c]{7em}{\includegraphics[width=6.7em]{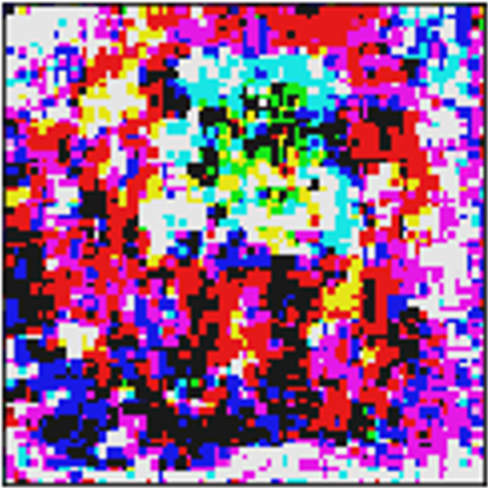}} 
  \\
 \parbox[c]{7em}{\includegraphics[width=6.7em]{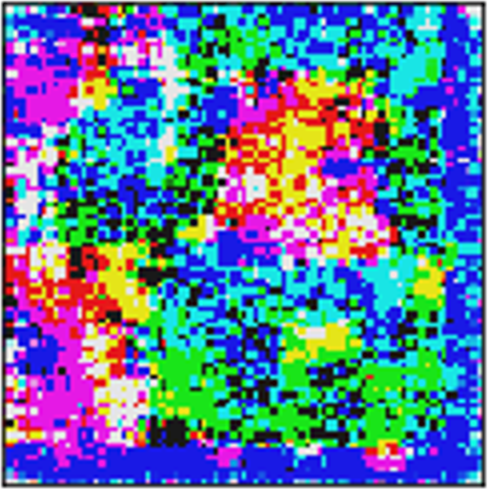}}
 \\
\end{tabular}
\end{tabular}
\end{tabular}
  \end{adjustbox}}
    \vspace*{-3mm}
    \caption{\small{Visualization of a randomly selected neuron's inverted input attribution maps (IAMs) under different meta-models. 
    %Here we randomly select a neuron of the four-layer convolutional models. 
    The first row shows the seed images. 
    The second-fourth rows show  IAMs corresponding to models trained by MAML, {\RMAMLa}, and {\RMAMLb}, respectively. Except  MAML, {\RMAMLa} and {\RMAMLb} all catch  high-level features from the data.
}
  }
  \vspace*{-5mm}
  \label{vis_inout}
\end{wrapfigure}
Spurred by experiment results in Figure\,\ref{comp_in_out}, 
we hypothesize that the promotion of robustness in meta-update \textit{alone} (i.e. {\RMAMLb}) is already sufficient to
offer \textit{robust representation}, over which   fine-tuned models  can preserve robustness to 
  downstream  tasks.
  In what follows, we   justify the above hypothesis  from   two perspectives: (i) explanation of learned neuron's representation and (ii)  resilience  of learnt robust meta-model to different fine-tuning schemes at the meta-testing phase.

\paragraph{(i) Learned signature of neuron's representation}
It is recently shown in \citep{engstrom2019adversarial} that  a robust model exhibits perceptually-aligned neuron activation maps, which  are not present if the model lacks adversarial robustness. 
To uncover such a  signature of robustness, a feature inversion technique \citep{engstrom2019adversarial} is applied to finding   an  inverted \textit{input attribution map} (IAM)  that maximizes  neuron's activation. Based on that, we examine if {\RMAMLa} and {\RMAMLb} can similarly   generate    explainable   inverted images from the learned neuron's representation. 
%than the standard MAML without imposing any robust regularization. 
We refer readers to Appendix \ref{ssec2} for more details on feature inversion from neuron's activation. 

\begin{wraptable}{r}{5.5cm}
\vspace*{-5mm}
\caption{\small{
Comparison of different strategies in meta-testing on {\RMAMLb}: %(10 steps PGD attack, $\epsilon=2$): 
(a) standard fine-tuning (S-FT), (b) adversarial fine-tuning (A-FT). }}
\begin{center}
\small
\label{acc_ft}
\resizebox{0.28\textwidth}{!}{
\begin{tabular}{l||c|c}
\hline
\hline
 & S-FT & A-FT \\
\hline 
SA &  40.9\% & 39.6\% \\

RA &  22.9\% & 23.5\% \\
\hline
\hline
\end{tabular}}
\end{center}
\vspace*{-5mm}
\end{wraptable}

In our experiment, we indeed find that both {\RMAMLa}  and  {\RMAMLb} yield similar IAMs inverted from neuron's activation   at different input examples, as plotted in Figure\,\ref{vis_inout}.   More intriguingly, the learnt IAMs characterize the contour of objects existed in input images, and accompanied by the learnt high-level features, e.g., colors. In contrast, the IAMs of MAML lack such an interpretability. The observations from the interpretability of neurons' representation justify why {\RMAMLb} is as effective  as {\RMAMLa} and why MAML does not preserve robustness.

% \SL{[a. PGD-attack (10 steps, $\epsilon = 2$) inner solver during training.
%  1-shot setting, plot figure, figure (a) SA \& RA (robust accuracy) vs. PGD attack steps for three methods Figure (b) SA \& RA vs. evaluation $\epsilon$ for three methods.
% ]}

% \SL{only robust meta-model is resilient to fine-tuning strategies during meta-testing.}

%In the aforementioned setting, 
%(different from fine-tuning scheme used during meta-training) 
% \SL{[Explanation maps: 
% Why it works. 
% ]}

% \SL{[Difference to existing works.Provide explanation from feature map.
% ]}

\paragraph{(ii) Robust meta-update provides robustness adaptation without additional adversarial fine-tuning at meta-testing}

Meta-testing includes only the fine-tuning stage. Therefore, we need to explore if standard fine-tuning is enough to maintain the robustness. Suppose that {\RMAMLb}  is  adopted as the meta-training method to solve problem \eqref{eq: prob_MAML_robust},  we then ask if robustness-regularized meta-testing strategy 
can improve the robustness of fine-tuned model at downstream tasks.
%to the learnt   meta-model can enhance robustness adaptation at the testing phase. 
Surprisingly, we find that making an additional effort
to adversarially fine-tune the meta-model (trained by  {\RMAMLb})
during testing does \textit{not}
%\textit{no longer} 
provide an
  obvious robustness improvement over the
  standard fine-tuning scheme during testing
%at the meta-testing phase
(Table\,\ref{acc_ft}).  
This consistently implies that  robust meta-update ({\RMAMLb}) is sufficient to render intrinsic robustness in its learnt meta-model regardless of fine-tuning strategies used at meta-testing. Figure\,\ref{vis_fine_tune} in Appendix \ref{ssec3} provides evidence that the visualization difference is small between before standard fine-tuning and after standard fine-tuning. 
%\Lily{This paragraph is less clear to me: 1. not sure what it means the meta-update is "resilient" to fine-tuning; 2. is "adversarially fine-fune the meta-model" here refers to setting $\gamma_{\textrm{in}} \neq 0$?}

%In the aforementioned setting, 
%(different from fine-tuning scheme used during meta-training) 
% \SL{[Explanation maps: 
% Why it works. 
% ]}

% \SL{[Difference to existing works.Provide explanation from feature map.
% ]}

\paragraph{Adversarial querying (AQ)  \citep{goldblum2019adversarially}: A special case of {\RMAMLb}}
The recent work \citep{goldblum2019adversarially}
developed  AQ to improve
 adversarial robustness in few-shot learning.  
 AQ can  be regarded as a special case of {\RMAMLb} with  $\gamma_{\mathrm{in}} = 0$ but setting   $\gamma_{\mathrm{out}} = \infty $ in \eqref{eq: prob_MAML_robust}.
That is, the meta-update is overridden by the AT regularization. We find that AQ  yields about $2\%$ RA improvement  over {\RMAMLb}, which uses $\gamma_{\mathrm{out}} = 0.2$ in \eqref{eq: prob_MAML_robust}. However, AQ leads to $11\%$   degradation in SA, and thus makes a much poorer robustness-accuracy tradeoff than our proposed {\RMAMLb}. 
We refer readers to Table~\ref{efficiency_maml} for comparison of the proposed {\RMAMLb} with other training baselines. 
Most importantly, different from \citep{goldblum2019adversarially}, we  provide insights on why   {\RMAMLb} is effective in promoting adversarial robustness from meta-update to fine-tuning.

\section{Computationally-Efficient Robustness-Regularized MAML}
\label{sec: where}
In this section, we study if the proposed {\RMAMLb} can further be improved  to ease of optimization
given the two computation difficulties in  \eqref{eq: prob_MAML_robust}:
(a) bi-leveled  meta-learning,  and (b) the need of  inner maximization  to find the worst-case  robust  regularization.
To tackle either problem alone, there have been efficient solution methods  proposed  recently. In \citep{raghu2019rapid}, an almost-no-inner-loop (ANIL) fine-tuning strategy was proposed, where fine-tuning is only applied to the task-specific classification head following a frozen representation network inherited from the meta-model.
Moreover, in \citep{Wong2020Fast}, a fast gradient sign method (FGSM) based attack generator was leveraged to improve the efficiency of AT  without losing its adversarial robustness.  %performance. 
Motivated by \citep{raghu2019rapid,Wong2020Fast}, we ask if  integrating {\RMAMLb}  with  ANIL and/or FGSM  can improve the training efficiency but preserves the robustness and generalization-ability of a meta-model learnt from  {\RMAMLb}.

% the representation block of a model is fixed during fine-tuning  but just 

% \SL{a) attack generation b) partial fine-tuning }

\mycomment{
Note that either MAML or adversarial training is computationally expensive. In this section, we propose an efficient robust MAML based on the Fast Attack Generator (FAG) \citep{Wong2020Fast}. The key idea is to combine the FGSM with random initialization of the perturbation. This method only requires one gradient step to generate adversarial examples and has proven effective in standard adversarial training. Table~\ref{FAG_PGD} compares the accuracy between MAML with Fast Attack Generator (MAML-FAG) and MAML with Projected Gradient Descent (MAML-PGD). The results show that MAML-FAG is as effective as PGD-based training in MAML. We also compare the time cost of different methods under 1GPU-Tesla V100 in Table~\ref{time_methods}. MAML+FAG (MU) has the shortest time compared to other adversarial training strategies. We want to emphasize that FAG is not limited to supervised learning loss, but can be extended to unsupervised learning loss. We will discuss more in the next section.
}

% (ANIP) Another efficient
% \SL{[no inner loop. missing experiments.]}

\paragraph{{\RMAMLb} meets ANIL and FGSM}
% \eqref{eq: prob_MAML_robust},
We decompose 
 the meta-model $\mathbf w = [{\modelR}, {\modelC}]$ into two parts: representation encoding network ${\modelR}$ and classification head ${\modelC}$. In 
{\RMAMLb}, namely, \eqref{eq: prob_MAML_robust} with $\gamma_{\mathrm{in}} = 0$, 
 ANIL  suggests to only fine-tune $\mathbf w_{\mathrm{c}}$  over a specific task $\task_i$. This leads to   
\begin{align}\label{eq: ANIL}
    \mathbf w_{\mathrm{c},i}^{\prime}
    = \argmin_{ \mathbf w_{\mathrm{c},i} }  
\ell_i ( \mathbf w_{\mathrm{c},i}, \mathbf w_{\mathrm{r}} ; \din_i, \mathbf w) , ~\text{with} ~ \mathbf w_{\mathrm{r},i}^{\prime} = \mathbf w_{\mathrm{r}}.\tag{ANIL}
\end{align}
In \ref{eq: ANIL}, the initialized representation network $\mathbf w_{\mathrm{r}}$ keeps intact during task-specific fine-tuning, which thus saves the computation cost. Furthermore, if FGSM is used  in {\RMAMLb}, then the robustness regularizer $\mathcal R$ defined in \eqref{eq: prob_robust} reduces to
\begin{align}\label{eq: FGSM}
    \mathcal R (\mathbf w; \mathcal D) 
    = \mathbb E_{(\mathbf x, y) \in \mathcal D} [  
 g( \mathbf w; \mathbf x+ \boldsymbol \delta^*(\mathbf x),  y ) ], \quad \boldsymbol \delta^*(\mathbf x) = \boldsymbol \delta_0 + \epsilon \nabla_{\mathbf x} g(\mathbf w; \mathbf x, y),
 \tag{FGSM}
\end{align}
where $\boldsymbol \delta_0$ is an initial point  randomly 
drawn from a uniform distribution over the interval $[-\epsilon, \epsilon]$. Note that in the original implementation of robust regularization $\mathcal R$, a multi-step projected gradient ascent (PGA) is typically used to optimize the sample-wise adversarial perturbation $\boldsymbol \delta (\mathbf w)$. By contrast, 
 \ref{eq: FGSM} only uses one-step PGA in attack generation and thus improves the computation efficiency. 
 
\begin{wraptable}{r}{80mm}
\vspace*{-5mm}
\caption{\small{
Performance of computation-efficient alternatives of  {\RMAMLb} in SA, RA and computation time per epoch (in minutes). 
%[Did you ever try {\RMAMLbAnilFgsm}? What is the performance?].
}}
\begin{center}
\small
\label{efficiency_maml}
\resizebox{0.57\textwidth}{!}{
\begin{tabular}{l||c|c|c}
\hline
\hline
& SA & RA &Time\\
\hline 
MAML & 43.6\% & 3.17\% & 42min \\
\hline 
AQ \citep{goldblum2019adversarially} & 29.6\% & 24.9\% & 52min \\
\hline
{\RMAMLb}& 40.9\% & 22.9\% &  54min \\
\hline
{\RMAMLbAnil} & 37.46\% & 22.7\% & 36min \\
\hline
{\RMAMLbFgsm} & 40.82\% & 23.04\% & 44min \\
\hline
\hline
\end{tabular}}
\end{center}
\vspace*{-5mm}
\end{wraptable}
In Table\,\ref{efficiency_maml}, we study   two computationally-light alternatives of {\RMAMLb},
{\RMAMLb} with \ref{eq: ANIL} ({\RMAMLbAnil}) and 
{\RMAMLb} with \ref{eq: FGSM} ({\RMAMLbFgsm}). 
% and 
% {\RMAMLb} with both \ref{eq: ANIL} and \ref{eq: FGSM} ({\RMAMLbAnilFgsm}).
Compared to {\RMAMLb},
we find that  although
{\RMAMLbFgsm} takes less computation time, it yields even better RA with slightly worse   SA. 
By contrast, {\RMAMLbAnil} yields the least computation cost but the worst SA and RA. For comparison, 
we also present the performance of the adversarial meta-learning baseline AQ \citep{goldblum2019adversarially}. As we can see,
AQ promotes the adversarial robustness  at the cost of a significant SA drop, e.g., $7.56\%$ worse than   {\RMAMLbAnil}.
Overall, the application of  \ref{eq: FGSM}  to {\RMAMLb} provides the most graceful tradeoff between the computation cost and the standard and robust accuracies. In the rest of the paper, unless specified otherwise we will use \ref{eq: FGSM}   in {\RMAMLb}.

% and the simplest implementation of {\RMAMLb}, 
% {\RMAMLbAnilFgsm}, which combines {\RMAMLb} with both \ref{eq: ANIL}  and \ref{eq: FGSM}

 \mycomment{
\textbf{Baseline Methods} In this paper, we will compare our methods to four baseline methods: (1) MAML (2) Adversarial Querying (AQ) \citep{goldblum2019adversarially} (3) Standard training + Meta-testing (4) Standard robust training + Meta-testing. %AQ applies the PGD attack to generate adversarial query data and only use the adversarial data in meta-update. 
For method (3), we train a model with standard training using the training data (60 classes) and fine-tune the model with few-shot learning using the test data (20 classes). Method (4) applies robust training \citep{madry2017towards} in meta-training and fine-tune the model using the test data under the few-shot learning framework.
}

% \begin{table}[h]
% \caption{Comparison of different metheds in terms of efficiency and accuracy. Compared to {\RMAMLb}-PGA, {\RMAMLbFgsm} and {\RMAMLbAnil} can obtain similar accuracy with faster speed.}
% \small
% \label{efficiency_maml}
% \begin{center}
% \resizebox{0.7\textwidth}{!}{
% \begin{tabular}{l||c|c|c}
% \hline
% \hline
% & SA & RA & Computation times (500 batches)\\
% \hline 
% MAML & 43.6\% & 3.17\% & 1060s \\
% \hline 
% {\RMAMLbFgsm} & 40.82\% & 23.04\% & 1100s \\
% \hline
% {\RMAMLb}-PGA & 40.9\% & 22.9\% &  1360s \\
% \hline
% AQ & 29.6\% & 24.9\% & 1300s \\
% \hline
% {\RMAMLbAnil} & 37.46\% & 22.7\% & 900s \\
% \hline
% \hline
% \end{tabular}}
% \end{center}
% \end{table}

\section{Semi-Supervised Robustness-Promoting MAML}

Given our previous solutions to \textit{when} (Sec.\,\ref{sec: when})  and \textit{how}  (Sec.\,\ref{sec: where})
a robust regularization could effectively be promoted in few-shot learning, we next ask: Is it possible to further improve our proposal {\RMAMLb} by leveraging \textit{unlabeled data}?
Such a question is motivated from two aspects.
First, the use of unlabeled data augmentation could be  a key momentum to improve the robustness-accuracy tradeoff \citep{carmon2019unlabeled,stanforth2019labels}. Second, the recent success in self-supervised contrastive representation learning \citep{chen2020simple,he2020momentum} demonstrates the power of multi-view (unlabeled) data augmentation to acquire  discriminative and generalizable visual representations, which can   guide down-stream supervised learning. 
%for downstream  
In what follows, we propose an extension of {\RMAMLb} applicable to  semi-supervised learning with unlabeled data augmentation.

\paragraph{{\RMAMLb} with TRADES regularization.}
We recall from \eqref{eq: prob_robust} that the robust regularization   $\mathcal R$ can also be specified by TRADES \citep{zhang2019theoretically}, which relies only on the prediction logits of benign and adversarial examples (rather than the training label), and thus
lends itself to the application of unlabeled data. 
Spurred by that, we propose {\RMAMLbTr}, which is a variant of {\RMAMLb} using the unlabeled data augmented TRADES regularization.
To perform data augmentation in experiments, we follow \citep{carmon2019unlabeled} to  mine  additional (unlabeled) data with the same amount of   MiniImagenet data from 
the original ImageNet data set.
For clarity, we call  {\RMAMLb} using TRADES or AT regularization (but without unlabeled data augmentation)    {\RMAMLb(TRADES)} or {\RMAMLb(AT)}.

\begin{wrapfigure}{r}{60mm}
\vspace*{-0.2in}
\centerline{
\includegraphics[width=.475\textwidth,height=!]{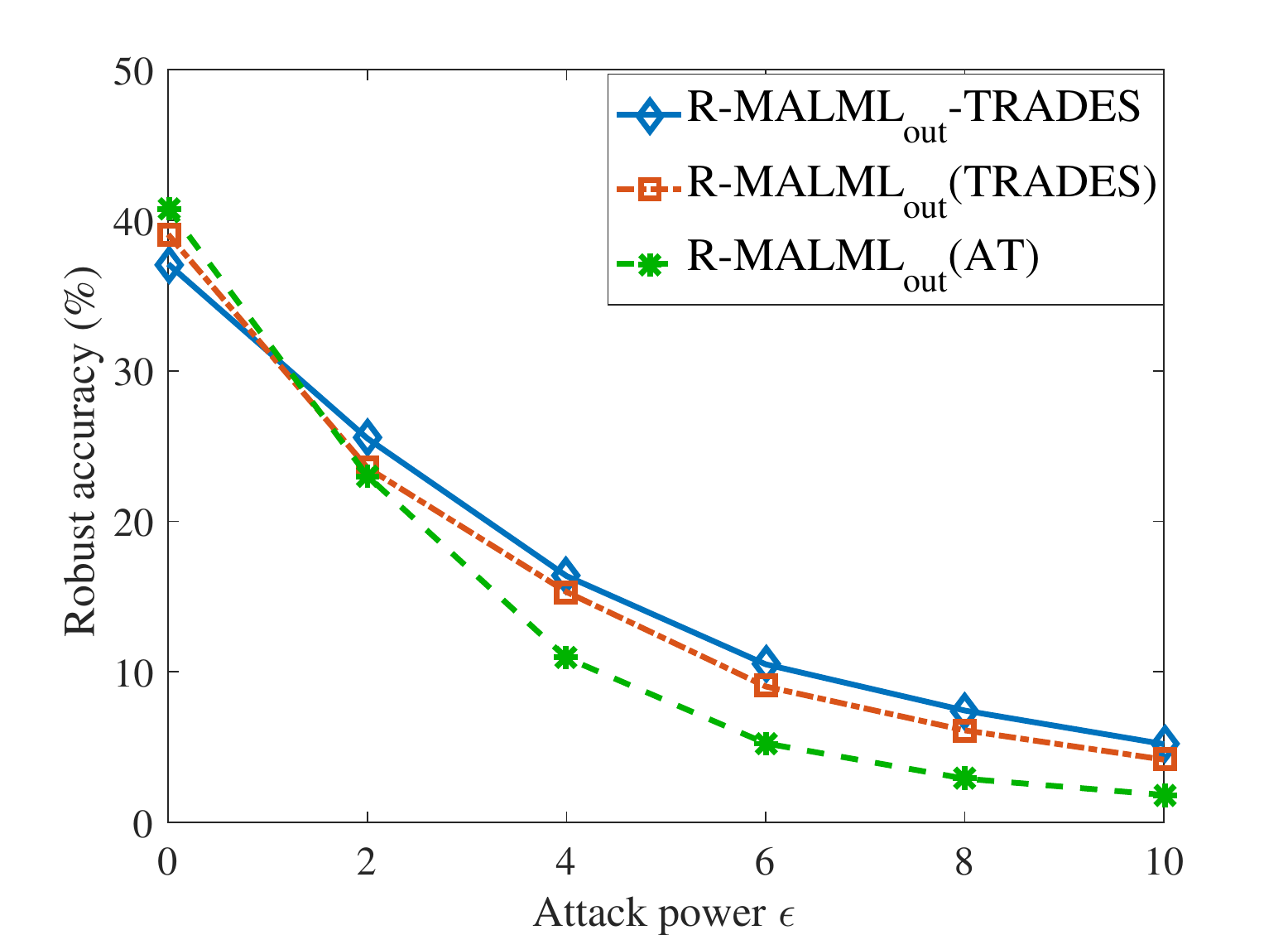}
}
\vspace*{-0.05in}
\caption{\small{{
RA versus (testing-phase) PGD attacks at different values of perturbation strength $\epsilon$. Here the robust  models are trained by different variants of {\RMAMLb}, including {\RMAMLbTr} (with unlabeled data augmentation), {\RMAMLb} using AT regularization but no   data  augmentation ({\RMAMLb}(AT)), and {\RMAMLb} using TRADES regularization but no data  augmentation ({\RMAMLb}(TRADES)).}
%when the attack power increases. MAML-TRADES-SSL obtains stronger defense when the attack power increases.
}}
  \label{adv_atkpower}
  \vspace*{-2mm}
\end{wrapfigure}
We find that with the help of unlabeled data, {\RMAMLbTr} improves the accuracy-robustness tradeoff over its supervised counterpart  {\RMAMLb} using either AT or TRADES regularization
(Figure~\ref{adv_atkpower}).
Compared to {\RMAMLb}, {\RMAMLbTr} yields consistently better RA against different attack strength  $\epsilon \in \{2,\ldots,10 \}/255$ during testing. Interestingly, the improvement becomes more significant as $\epsilon$ increases. 
As $\epsilon = 0$, RA is  equivalent to SA, and we observe that the superior   performance of  {\RMAMLbTr} in  RA bears a slight degradation in SA compared to 
{\RMAMLb(TRADES)} and {\RMAMLb(AT)}, which indicates the robustness-accuracy tradeoff. Figure\,\ref{comp_dif_eps} in Appendix \ref{ssec5} provides an additional evidence that {\RMAMLbTr} has the ability to defend stronger attacks than {\RMAMLb}, and proper unlabeled data augmentation can further improve the accuracy-robustness tradeoff in MAML.

% Figure~\ref{adv_atkpower} shows the adversarial accuracy of different models when the attack power increases. {\RMAMLb} and MAML-TRADES-SSL are trained using $\epsilon = 2$ and $\epsilon = 4$, respectively. Compared to {\RMAMLb}, MAML-TRADES-SSL reaches higher robust accuracy when the attack power is the same as the attack generation during the training and obtains stronger defense when the attack power increases. The results indicate that proper selecting unlabeled data can help to improve the robustness.

\paragraph{{\RMAMLb} with contrastive learning (CL).}
To improve adversarial robustness, many   works, e.g., \citep{pang2019improving,sankaranarayanan2017regularizing}, also suggest that 
it is important to
encourage robust semantic features that locally cluster according to class, namely, ensuring
that features of samples in the same class will lie close to each other and away from those of different
classes.
The above suggestion  aligns  with the goals of contrastive learning (CL) \citep{chen2020simple,wang2020understanding}, which promotes
(a) alignment (closeness) of features
from positive data pairs, and (b) uniformity of feature distribution. 
Thus, we develop {\RMAMLbCL} by integrating   {\RMAMLb} with
CL.

Prior to defining {\RMAMLbCL}, we first introduce
 CL  and refer   readers to \citep{chen2020simple} for   details. 
% Let  $p(\cdot)$ be the data distribution over $\mathbb R^{n}$ and 
% $p^+(\cdot, \cdot)$ be  
% the distribution of positive pairs over $\mathbb R^{n  } \times \mathbb R^n$.
Given a data sample $\mathbf x$, CL utilizes its positive counterpart $\mathbf x^+$   given by a certain data transformation $t$, e.g., cropping and resizing, cut-out, and rotation, $\mathbf x^+ = t(\mathbf x)$.  The data pair $(\mathbf x, t (\mathbf x^{\prime}))$ is then positive if 
$\mathbf x = \mathbf x^{\prime}$, and negative otherwise. 
% pair becomes  if $\mathbf x \neq \mathbf x^{\prime}$, where $t$ and $t^{\prime}$
The \textit{contrastive loss} is defined by 
{\small\begin{align} %\label{eq: CL_loss}
    \ell_{\mathrm{CL}}(\mathbf w_{\mathrm{c}}; p^+) =  \mathbb E_{(\mathbf x, \mathbf x^+ ) \sim p^+} 
    \left [
    - \log  \frac{e^{  { \mathbf r ( \mathbf x ; \mathbf w_{\mathrm{c}})^T 
    \mathbf r( \mathbf x^+ ; \mathbf w_{\mathrm{c}}) }/{\tau} 
    }}{
    e^{  { \mathbf r ( \mathbf x ; \mathbf w_{\mathrm{c}})^T 
    \mathbf r( \mathbf x^+ ;\mathbf w_{\mathrm{c}}) }/{\tau} 
    } +  \sum_{
            \mathbf x^{-} \sim p,  
        (\mathbf x, \mathbf x^-) \notin p^+
    } \left [ 
    e^{  { \mathbf r ( \mathbf x ;\mathbf w_{\mathrm{c}})^T 
    \mathbf r( \mathbf x^- ;\mathbf w_{\mathrm{c}}) }/{\tau} 
    }
    \right ]
    }
    \right ], \nonumber
\end{align}}%
where $\mathbf x \sim p$
denotes the data distribution, $p^+(\cdot, \cdot)$    
 is the distribution of positive pairs,  
$\mathbf r(\mathbf x; \mathbf w_{\mathrm{c}})$ is the encoded representation of $\mathbf x$ extracted from the representation network $\mathbf w_{\mathrm{c}}$, 
and $\tau >0$ is a temperature parameter. 
The contrastive loss minimizes  
the distance of a positive pair among many negative pairs, namely, learns network representation with
  instance-wise discriminative power.

According to CL, 
we then  augment the data used to train {\RMAMLb} with their transformed   counterparts. In addition, the  \textit{adversarial examples} generated during   robust regularization can also be used as \textit{additional  views} of the original data, which  in turn advance CL.
Formally, we modify {\RMAMLb}, given by \eqref{eq: prob_MAML_robust} with $\gamma_{\mathrm{in}} = 0$,    as 
\begin{align}\label{eq: prob_MAML_robust_CL}
   % \hspace*{-0.05in}
    \begin{array}{ll}
\displaystyle \minimize_{\mathbf w}         & \frac{1}{N} \sum_{i=1}^N \left [  \ell_i^{\prime}( \mathbf w_i^{\prime}; \dout_i ) + \gamma_{\mathrm{out}}  \mathcal R_i(\mathbf w_i^{\prime}; \dout_i) + \gamma_{\mathrm{CL}}  \ell_{\mathrm{CL}}(\mathbf w_{\mathrm{c},i}^{\prime}; p_{i}^{+ } \cup p_{i}^{\mathrm{adv}}   )  \right ]  \\
      \st   &   \mathbf w_i^{\prime} = \argmin_{\mathbf w_i} 
\ell_i (\mathbf w_i; \din_i, \mathbf w)  , ~\forall i \in [N],
    \end{array}
   % \hspace*{-0.15in}
\end{align}
where $\gamma_{\mathrm{CL}} > 0$ is a regularization parameter associated with the contrastive loss, $p_{i}^{+ } \cup p_{i}^{\mathrm{adv}}$ represents the distribution of positive data pairs constructed by the standard    and adversarial views of $\mathcal D^{\prime}$,  and 
 $\mathbf w_{\mathrm{c},i}^{\prime}$ denotes the representation block of the model $\mathbf w_i^{\prime}$.

\begin{wraptable}{r}{77mm}
\vspace*{-5mm}
\caption{\small{
SA/RA performance of {\RMAMLbCL} versus other variants of proposed {\RMAMLb} and baselines.
}}
\vspace*{-1mm}
\begin{center}
\small
\label{tab: summary_CL}
\resizebox{0.55\textwidth}{!}{
\begin{tabular}{l||c|c}
\hline
\hline
 & SA & RA \\
\hline
MAML & \bf{43.6\%} & 3.17\%    \\
\hline
AQ \citep{goldblum2019adversarially} & 29.6\% & 24.9\%  \\
\hline
% Standard training + Meta-testing & 29.74\% & 3.51\%  \\
% \hline
% Standard robust training + Meta-testing & 28.22\% & 19.02\% \\
% \hline
{\RMAMLb}(AT) (ours) & 40.82\% & 23.04\%   \\
\hline
{\RMAMLb}(TRADES) (ours) & 39.06\% & 23.56\%  \\
\hline
{\RMAMLb}-TRADES (ours) & 37.1\% & 25.51\% \\
\hline
{\RMAMLbCL} (ours) & 38.60\% & \bf{26.81\%} \\
\hline
\hline
\end{tabular}}
\end{center}
\vspace*{-1mm}
\end{wraptable}
In Table\,\ref{tab: summary_CL}, we compare the SA/RA performance of {\RMAMLbCL} with that of  
previously-suggested $3$ variants of {\RMAMLb} including the versions 
{\RMAMLb}(AT) and  {\RMAMLb}(TRADES) without using unlabeled data, 
and the version with unlabeled data {\RMAMLbTr}, as well as $2$ baseline methods including
standard MAML and adversarial querying (AQ) in few-shot learning  \citep{goldblum2019adversarially}. 
Note that we specify $\mathcal R_i$ in \eqref{eq: prob_MAML_robust_CL} as TRADES regularization for {\RMAMLbCL}.
We find that {\RMAMLbCL} yields   the best RA among all meta-learning methods, and    
improves SA over  {\RMAMLbTr}. 
In particular, the comparison with AQ shows that {\RMAMLbCL} leads to $9\%$ improvement in SA and $1.9\%$ improvement in RA.

% but also the best SA among all robustness regularized MAML methods.
% %{\RMAMLb}(AT) (Sec.\,\ref{sec: when}), {\RMAMLb}(TRADES) (Sec.\,\ref{sec:}) (Sec.\,\ref{sec: when})

% Tabel~\ref{1shot_ourmethods} shows the performances of our methods in 1-shot 5-way scenario. One can see that the proposed methods can all reach high robust accuracy as while as high standard accuracy compared to the baseline methods. In particular, MAML with semi-supervised learning and contrastive learning (MAML-TRADES-SSL-CL) reaches the highest adversarial accuracy - $26.81\%$.

% \subsection{Meta Learning of Robust Representation With Contrastive Learning (CL)}
% \label{meta_cl}

% %\subsubsection{Robust Regularization Loss with Labeled Data}

% %subsubsection{Robust Regularization Loss with Unlabeled Data}

% \subsection{Preliminaries of contrastive learning (CL)}

% \subsection{MAML augmented with CL}

% \SL{Missing implementation details.} {\color{blue} I moved this part to Appendix A}

% \subsection{Experimental Setup}

% \textbf{Evaluation Metrics} There are three evaluation metrics to measure the performance: (1) clean accuracy (CA) - classification accuracy for clean test data (2) adversarial accuracy (AA) - the classification accuracy under a certain attack strength (3) conditional adversarial accuracy (CAA) - classification accuracy for the correct labeled clean data under a certain attack strength

% \subsection{Performances of our Methods}

\section{Additional Experiments}
\label{exp_note}

\begin{wraptable}{r}{85mm}
\vspace*{-5mm}
\caption{\small{Summary of baseline performance in SA and TA.}
}
\begin{center}
\small
\label{comp_baseline}
\resizebox{0.6\textwidth}{!}{
\begin{tabular}{l||c|c}
\hline
\hline
 & SA & RA \\
\hline
MAML (FSL) & \bf{43.6\%} & 3.17\%    \\
\hline
AQ (FSL) \citep{goldblum2019adversarially} & 29.6\% & \bf{24.9\%}  \\
\hline
Supervised standard training (non-FSL) & 29.74\% & 3.51\%  \\
\hline
Supervised AT  (non-FSL)
& 28.22\% & 19.02\% \\
\hline
\hline
\end{tabular}}
\end{center}
\end{wraptable}

\paragraph{Key facts of our implementation.}
In the previous analysis, we consider 1-shot 5-way image classification tasks over
%in  under 
MiniImageNet  \citep{vinyals2016matching}. And we use  a four-layer convolutional neural network for few-shot learning (FSL). By default, we set the training attack strength $\epsilon=2$, $\gamma_{\mathrm{CL}}=0.1$, and set $\gamma_{\mathrm{out}}=5$ (TRADES), $\gamma_{\mathrm{out}}=0.2$ (AT) via a grid search. During meta-testing, a $10$-step PGD attack with attack strength $\epsilon = 2$ is used to evaluate RA of the learnt meta-model over $2400$ few-shot test tasks. We provide experiment details in Appendix\,\ref{ssec4}.

\paragraph{Summary of baselines.} 
We remark that in addition to MAML and AQ baselines, we also consider the other two baseline methods, supervised standard training over  the entire dataset (non-FSL setting), and supervised AT over the entire dataset (non-FSL setting); see a summary in Table~\ref{comp_baseline}.
The additional baselines demonstrate that robust adaptation in FSL is non-trivial as neither the supervised full AT or the full standard training can achieve satisfactory SA and RA. 
% provide the robustness adaptation performance of models normally trained over the entire dataset to few-shot test tasks.

\begin{table}[h]%{r}{95mm}
%\vspace*{-5mm}
\caption{
\small{SA/RA performance of our proposed methods on CIFAR-FS \citep{bertinetto2018meta}.}
}
\begin{center}
\small
\label{cifarfs_main}
\resizebox{0.7\textwidth}{!}{
\begin{tabular}{l||c|c|c|c}
\hline
\hline
& \multicolumn{2}{c|}{1-Shot 5-Way} & \multicolumn{2}{c}{5-Shot 5-Way}\\ [0.5ex] 
\hline
 & SA & RA & SA & RA \\
\hline
MAML & \bf{51.07\%} & 0.235\%  & \bf{67.2\%} & 0.225\%  \\
\hline
AQ \citep{goldblum2019adversarially} & 31.25\% & 26.34\% & 52.32\% & 33.96\%  \\
\hline
{\RMAMLb}(AT) (ours) & 39.76\% & 26.15\%  & 57.18\% & 32.62\% \\
\hline
{\RMAMLb}(TRADES) (ours) & 40.23\% & 27.45\%  & 57.46\% & 34.72\%   \\
\hline
{\RMAMLb}-TRADES (ours) & 40.59\% & 28.06\%  & 57.62\% & 34.76\%   \\
\hline
{\RMAMLbCL} (ours) & 41.25\% & \bf{29.33\%} & 57.95\% & \bf{35.30\%}  \\
\hline
\hline
\end{tabular}}
\end{center}
\end{table}

% \begin{wrapfigure}{r}{90mm}
% \vspace*{-0.2in}
% 	\begin{minipage}[t]{0.5\linewidth}
% 		\centering
% 		\includegraphics[trim=0 0 0 0,clip,width=.95\textwidth,height=!]{./Figs/omni_RA}
% 					%\vspace{-0.2in}
% 					{\scriptsize \center (a) RA}
% 	\end{minipage}%
% 	\begin{minipage}[t]{0.5\linewidth}
% 		\centering
% 		\includegraphics[trim=0 0.2 0 0,clip,width=.96\textwidth,height=!]{./Figs/omni_SA}
% 					%\vspace{-0.2in}
% 					{\scriptsize \center (b) SA}
% 	\end{minipage}
% 	\caption{\tcr{\small{Performance of {\RMAMLb}(TRADES) and AQ \citep{goldblum2019adversarially} on Omniglot versus number of classes in each task (from $5$ to $20$ ways): (a) RA. (b) SA.}} }
% 	\label{omniglot_comp_main}
% 	  \vspace*{-2mm}
% \end{wrapfigure}

\paragraph{Experiments on  Additional  model architecture, datasets and FSL setups.}
In Table\,\ref{1shot_resnet} of Appendix\,\ref{ssec5}, we provide additional experiments using ResNet18. In particular,
{\RMAMLbCL} leads to $13.94\%$ SA improvement and $1.42\%$ RA improvement over AQ. 
%The results indicate that our proposed methods do not rely on the specific model architecture.
% }
% %\paragraph{Experiments on various datasets and FSL configurations.}
% \tcr{
We also test our methods on CIFAR-FS \citep{bertinetto2018meta} and Omniglot \citep{lake2015human}, and provide the results in Table\,\ref{cifarfs_main} and Figure\,\ref{omniglot_comp}, respectively (more details can be viewed in Appendix\,\ref{ssec6} and Appendix\,\ref{ssec7}). The results show that our methods perform well on various datasets and outperform the baseline methods. On CIFAR-FS, we study 1-Shot 5-Way and 5-Shot 5-Way settings. As shown in Table\,\ref{cifarfs_main}, the use of unlabeled data augmentation ({\RMAMLbCL}) on CIFAR-FS can provide $10\%$ (or $5.6\%$) SA improvement and $3\%$ (or $1.3\%$) RA improvement over AQ under the 1-Shot 5-Way (or 5-Shot 5-Way) setting. Furthermore, we conduct experiments in other FSL setups. On Omniglot, we compare {\RMAMLb}(TRADES) to AQ \citep{goldblum2019adversarially} in the 1-shot (5, 10, 15, 20)-Way settings. Figure\,\ref{omniglot_comp} shows that {\RMAMLb}(TRADES) can always obtain better performance than AQ when the number of classes in each task varies.

\section{Conclusion}
In this paper, we study the problem of adversarial robustness in MAML. Beyond directly integrating MAML with robust training, we show and explain when a robust regularization should be promoted in MAML. We find that robustifying the meta-update stage via fast attack generation method is sufficient to achieve fast robustness  adaptation without losing generalization and computation efficiency in general. 
To further improve our proposal, we for the first time study how unlabeled data help robust MAML. In particular, we propose using contrastive representation learning to acquire improved generalization and robustness simultaneously. 
% We first look at where to implement adversarial training, find that it is sufficient to incorporate adversarial training in the meta-update stage, and implement the standard fine-tuning in the meta-testing. We also provide a visualization insight to support our findings. Then we propose a general framework that allows efficient adversarial training, introduces the semi-supervised learning, and easy fine-tuning. We for the first time show that introducing the contrastive learning in our objective can improve the standard accuracy/robust accuracy performance. Finally, 
Extensive experiments are provided to demonstrate the effectiveness of our approach and justify our  insights on the adversarial robustness of MAML. In the future, we plan to establish the convergence rate analysis of robustness-aware MAML by leveraging bi-level and min-max optimization theories. %We also plan to 

% \tcr{In terms of theoretical analysis, we remark that both MAML and adversarial training (AT) have very limited progress in theory. The reason is that both MAML and AT are in the form of quite challenging optimization problems, where the former is given by a bi-level optimization problem and the latter is given by a min-max optimization problem. The existing theoretical analysis for MAML and AT \citep{hong2020two,fallah2020convergence,gao2019convergence} is restricted to the convergence rate of bi-level and min-max optimization algorithms under strong assumptions, e.g. convexity required for the inner-level optimization problem. What is more, our setting becomes more difficult to analyze as AT (min-max optimization) is embedded in MAML (bi-level optimization). Thus, although having theoretical analysis is attractive, it is much beyond the scope of our work.}
% % the effectiveness of our methods. We believe this work delivers valuable lessons and design principles for efficient robustness adaptation in the meta-learning framework.

\section*{Acknowledgement}
This work was supported by the Rensselaer-IBM AI Research Collaboration (\url{http://airc.rpi.edu}), part of the IBM AI Horizons Network (\url{http://ibm.biz/AIHorizons}).

\bibliography{reference,refs_adv}
%\bibliography{iclr2021_conference}
\bibliographystyle{iclr2021_conference}

\newpage
\clearpage

\setcounter{section}{0}

\section*{Supplementary Material}

\setcounter{section}{0}
\setcounter{figure}{0}
\makeatletter 
\renewcommand{\thefigure}{S\@arabic\c@figure}
\makeatother
\setcounter{table}{0}
\renewcommand{\thetable}{S\arabic{table}}
\setcounter{algorithm}{0}
\renewcommand{\thealgorithm}{S\arabic{algorithm}}
\setcounter{equation}{0}
\renewcommand{\theequation}{S\arabic{equation}}

\section{Framework of {\RMAMLb}}\label{ssec1}

Algorithm \ref{rmaml} shows the framework of {\RMAMLb}. The initial inputs include model weights $\mathbf w$, distribution of the training tasks $p(\mathcal{T})$, and the step sizes $\alpha, \beta_1, \beta_2$, which correspond to fine-tuning, clean meta-update, adversarial meta-update. Each batch contains multiple tasks that are sampled from the $p(\mathcal{T})$. $K$ is the number of gradient updates in fine-tuning. The adapted parameter $\mathbf w_i^{(K)}$ is used to generate adversarial validation data $\hat{\din_i^{\prime}}$ from the clean validation data $\din_i^{\prime}$ and to compute the loss value $\mathcal R_i(\mathbf w_i^{(K)}; \hat{\din_i^{\prime}})$. The attack generator can be selected from Projected Gradient Descent \citep{madry2017towards}, Fast Gradient Sign Method \citep{goodfellow2014explaining}, etc. Here $\epsilon$ is used to control the attack strength in the training.

\begin{algorithm}[h]
\caption{{\RMAMLb}}
\label{rmaml}
\begin{algorithmic}[1]
\REQUIRE The initialization weights $\mathbf w$; Distribution over tasks $p(\mathcal{T})$; Step size parameters $\alpha, \beta_1, \beta_2$.
\WHILE{not done}
\STATE{Sample batch of tasks $\mathcal{T}_i \sim p(\mathcal{T})$ and separate data in $\mathcal{T}_i$ into $(\din_i, \din_i^{\prime})$}
\FORALL{$\mathcal{T}_i$}
\STATE{$\mathbf w_i^{(0)} := \mathbf w$}
\FOR{$k=1,2,\cdots,K$}
\STATE{$\mathbf w_i^{(k)} =  \mathbf w_i^{(k-1)} - \alpha  \nabla_{\mathbf w_i}  \ell_i( \mathbf w_i^{(k-1)} ; \din_i, \mathbf w )$}
\ENDFOR
\STATE{Using attack generator to generate adversarial validation data $\hat{\din_i^{\prime}}$ by maximizing adversarial loss $\mathcal R_i(\mathbf w_i^{(K)}; \hat{\din_i^{\prime}})$ with the constraint $\|\hat{\din_i^{\prime}} - \din_i^{\prime}\|_{\infty} \le \epsilon$}
\ENDFOR
\STATE{$\mathbf w := \mathbf w - \beta_1\nabla_{\mathbf w}\sum_{\mathcal{T}_i \sim p(\mathcal{T})}\ell_i( \mathbf w_i^{(K)} ; \din_i^{\prime}, \mathbf w ) - \beta_2 \gamma_{\mathrm{out}}\nabla_{\mathbf w}\sum_{\mathcal{T}_i \sim p(\mathcal{T})} \mathcal R_i(\mathbf w_i^{(K)}; \hat{\din_i^{\prime}}) $}
\ENDWHILE
\RETURN $\mathbf w$
\end{algorithmic}
\end{algorithm}

\section{Details of Learned Signature of Neuron's Activation}\label{ssec2}

By maximizing a single coordinate of the neuron activation vector $\mathbf r$ (the output before the fully-connected layer) with a perturbation in the input, the perturbation will show different behaviors between a robust model and a standard model \citep{engstrom2019adversarial}. To be more specific, the feature pattern is revealed in the input under a robust model, while a standard model does not have such behavior. The optimization problem can be mathematically written in the following form

\begin{align}\label{neural_act}
    \begin{array}{ll}
\displaystyle\maximize_{\boldsymbol{\delta}}         &  r_i(\mathbf x + \boldsymbol \delta) \\
     \st     &  - \mathbf x_j \le \boldsymbol{\delta}_j \le 255 - \mathbf x_j, 
    \end{array}
\end{align}

where $r_i$ denotes the $i$-th coordinate  of  neuron activation vector. $\boldsymbol \delta$ is the perturbation in the input. $\mathbf x_j$ is the $j$-th pixel of the image vector $\mathbf x$.

\section{Visualization of IAMs Before and After Fine-Tuning in Meta-Testing}\label{ssec3}

\begin{figure}[H]
  \centering
  \begin{adjustbox}{max width=0.6\textwidth }
    \begin{tabular}{@{\hskip 0.00in}c  @{\hskip 0.02in} c @{\hskip 0.02in} @{\hskip 0.02in} c }
%   & 
%   \begin{tabular}{@{\hskip 0.00in}c  @{\hskip 0.02in} c @{\hskip 0.02in} @{\hskip 0.02in} c }
% \centering\colorbox{lightgray}{\large \textbf{Neuron 1}}
% &
% \hspace*{0.4in}
% \centering\colorbox{lightgray}{\large \textbf{ Neuron 2}}
% \end{tabular}

% &
%   \begin{tabular}{@{\hskip 0.00in}c  @{\hskip 0.02in} c @{\hskip 0.02in} @{\hskip 0.02in} c }
% \centering\colorbox{lightgray}{\large \textbf{ Neuron 1}}
% &\hspace*{0.05in}
% \centering\colorbox{lightgray}{\large \textbf{ Neuron 2}}
% \end{tabular}

% \\

%   \begin{tabular}{@{\hskip 0.00in}c  @{\hskip 0.02in} c @{\hskip 0.02in}c @{\hskip 0.02in} c @{\hskip 0.02in} c  @{\hskip 0.02in}c @{\hskip 0.02in}c @{\hskip 0.02in} c @{\hskip 0.02in}c }
 \begin{tabular}{@{}c@{}}  
\vspace*{0.01in}\\
\rotatebox{90}{\parbox{9em}{\centering \normalsize \textbf{Seed Images}}}
% \vspace*{-0.05in}
 \\
% \rotatebox{90}{\parbox{9em}{\centering \normalsize \textbf{\begin{tabular}[c]{@{}c@{}}Recovered\\ images\\(Before-FT)  \end{tabular}  }}}
%  \\
\rotatebox{90}{\parbox{9em}{\centering \normalsize \textbf{\begin{tabular}[c]{@{}c@{}}IAMs\\(Before-FT)  \end{tabular}}}}
\\
% \rotatebox{90}{\parbox{9em}{\centering \normalsize \textbf{\begin{tabular}[c]{@{}c@{}}Recovered\\ images\\(After-FT)  \end{tabular} }}}
% \\
\rotatebox{90}{\parbox{9em}{\centering \normalsize \textbf{\begin{tabular}[c]{@{}c@{}}IAMs\\(After-FT)  \end{tabular} }}}
\\
\end{tabular} 
&
 \begin{tabular}{@{\hskip 0.02in}c@{\hskip 0.02in}c@{\hskip 0.02in}}
 \begin{tabular}{@{\hskip 0.02in}c@{\hskip 0.02in}c@{\hskip 0.02in}c@{\hskip 0.02in}c@{\hskip 0.02in}}
\begin{tabular}{@{\hskip 0.02in}c@{\hskip 0.02in}}
\\
 \parbox[c]{9em}{\includegraphics[width=9em]{./Figs/visclean/seedimgnet1}} 
%  \\
%  \parbox[c]{9em}{\includegraphics[width=9em]{./Figs/visclean/n483fagrec1}}
  \\
  \parbox[c]{9em}{\includegraphics[width=9em]{./Figs/visclean/n483fagpet1}}
  \\
%   \parbox[c]{9em}{\includegraphics[width=9em]{./Figs/visclean/n483fagaftrec1}}
%   \\
 \parbox[c]{9em}{\includegraphics[width=9em]{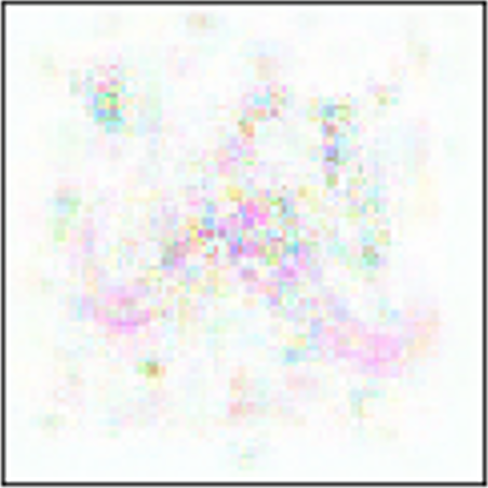}}
 \\
\end{tabular}

&
 \begin{tabular}{@{\hskip 0.02in}c@{\hskip 0.02in}}
\\
 \parbox[c]{9em}{\includegraphics[width=9em]{./Figs/visclean/seedimgnet2}} 
 \\
%  \parbox[c]{9em}{\includegraphics[width=9em]{./Figs/visclean/n483fagrec2}}
%   \\
  \parbox[c]{9em}{\includegraphics[width=9em]{./Figs/visclean/n483fagpet2}}
  \\
%   \parbox[c]{9em}{\includegraphics[width=9em]{./Figs/visclean/n483fagaftrec2}}
%   \\
 \parbox[c]{9em}{\includegraphics[width=9em]{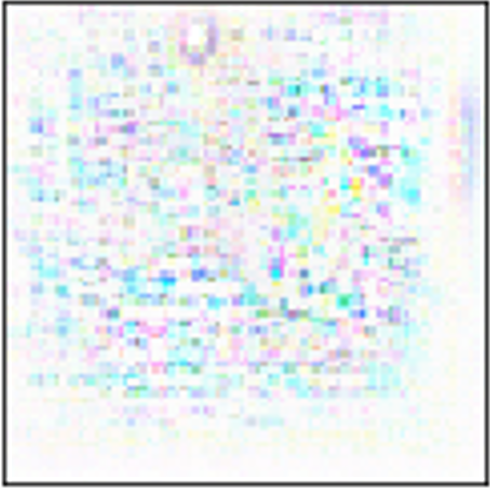}}
 \\
\end{tabular}
&
% \begin{tabular}{@{\hskip 0.02in}c@{\hskip 0.02in}c@{\hskip 0.02in}}
% \\
%  \parbox[c]{9em}{\includegraphics[width=9em]{./Figs/visclean/seedimgnet1}} 
%  \\
%  \parbox[c]{9em}{\includegraphics[width=9em]{./Figs/visclean/n683fagrec1}}
%   \\
%   \parbox[c]{9em}{\includegraphics[width=9em]{./Figs/visclean/n683fagpet1}}
%   \\
%   \parbox[c]{9em}{\includegraphics[width=9em]{./Figs/visclean/n683fagaftrec1}}
%   \\
%  \parbox[c]{9em}{\includegraphics[width=9em]{./Figs/visclean/n683fagaftpet1}}
% \\
% \end{tabular}
% &
%  \begin{tabular}{@{\hskip 0.02in}c@{\hskip 0.02in}}
% \\
%  \parbox[c]{9em}{\includegraphics[width=9em]{./Figs/visclean/seedimgnet2}} 
%  \\
%  \parbox[c]{9em}{\includegraphics[width=9em]{./Figs/visclean/n683fagrec2}}
%   \\
%   \parbox[c]{9em}{\includegraphics[width=9em]{./Figs/visclean/n683fagpet2}}
%   \\
%   \parbox[c]{9em}{\includegraphics[width=9em]{./Figs/visclean/n683fagaftrec2}}
%   \\
%  \parbox[c]{9em}{\includegraphics[width=9em]{./Figs/visclean/n683fagaftpet2}}
%  \\
% \end{tabular}
\end{tabular}
\end{tabular}
&
 \begin{tabular}{@{\hskip 0.02in}c@{\hskip 0.02in}}
 \begin{tabular}{@{\hskip 0.02in}c@{\hskip 0.02in}c@{\hskip 0.02in}c@{\hskip 0.02in}c@{\hskip 0.02in}}
 \begin{tabular}{@{\hskip 0.02in}c@{\hskip 0.02in}}
\\
 \parbox[c]{9em}{\includegraphics[width=9em]{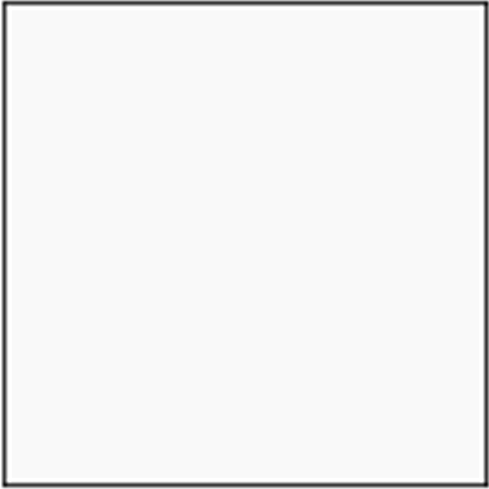}} 
 \\
%  \parbox[c]{9em}{\includegraphics[width=9em]{./Figs/visconst/nn483fagrec1}} 
%   \\
  \parbox[c]{9em}{\includegraphics[width=9em]{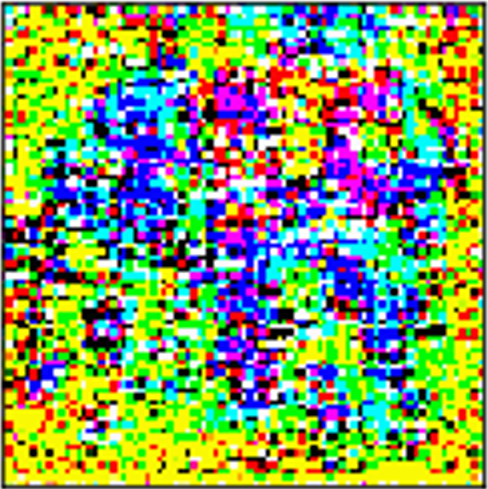}} 
  \\
%   \parbox[c]{9em}{\includegraphics[width=9em]{./Figs/visconst/nn483fagaftrec1}} 
%   \\
 \parbox[c]{9em}{\includegraphics[width=9em]{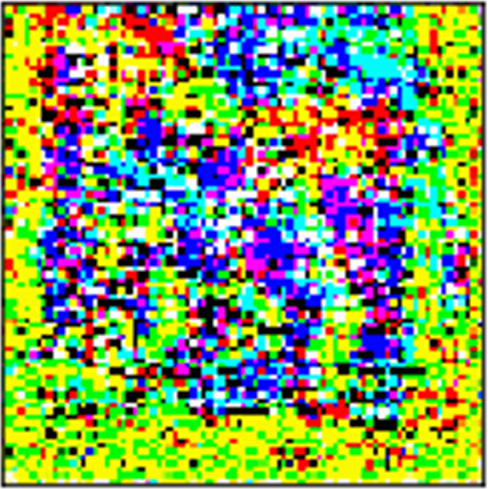}}
 \\

\end{tabular}
&
 \begin{tabular}{@{\hskip 0.02in}c@{\hskip 0.02in}}
\\
 \parbox[c]{9em}{\includegraphics[width=9em]{./Figs/visconst/seedconst2}} 
 \\
%  \parbox[c]{9em}{\includegraphics[width=9em]{./Figs/visconst/nn483fagrec2}} 
%   \\
  \parbox[c]{9em}{\includegraphics[width=9em]{./Figs/visconst/nn483fagpet2}} 
  \\
%   \parbox[c]{9em}{\includegraphics[width=9em]{./Figs/visconst/nn483fagaftrec2}} 
%   \\
 \parbox[c]{9em}{\includegraphics[width=9em]{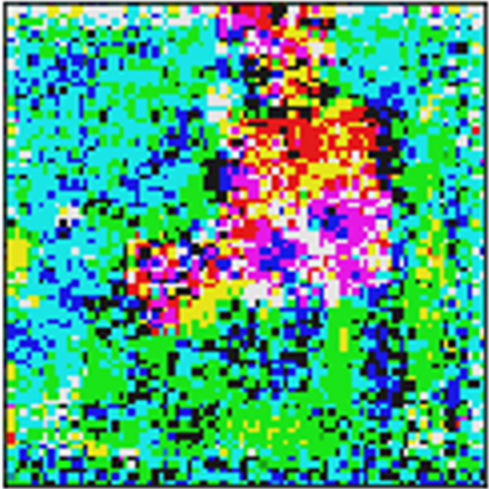}}
 \\
\end{tabular}

% &
%  \begin{tabular}{@{\hskip 0.02in}c@{\hskip 0.02in}}
% \\
%  \parbox[c]{9em}{\includegraphics[width=9em]{./Figs/visconst/seedconst1}} 
%  \\
%  \parbox[c]{9em}{\includegraphics[width=9em]{./Figs/visconst/nn683fagrec1}}
%   \\
%   \parbox[c]{9em}{\includegraphics[width=9em]{./Figs/visconst/nn683fagpet1}}
%   \\
%   \parbox[c]{9em}{\includegraphics[width=9em]{./Figs/visconst/nn683fagaftrec1}}
%   \\
%  \parbox[c]{9em}{\includegraphics[width=9em]{./Figs/visconst/nn683fagaftpet1}}
%  \\
% \end{tabular}
% &
% \begin{tabular}{@{\hskip 0.02in}c@{\hskip 0.02in}}
% \\
%  \parbox[c]{9em}{\includegraphics[width=9em]{./Figs/visconst/seedconst2}} 
%  \\
%  \parbox[c]{9em}{\includegraphics[width=9em]{./Figs/visconst/nn683fagrec2}}
%   \\
%   \parbox[c]{9em}{\includegraphics[width=9em]{./Figs/visconst/nn683fagpet2}}
%   \\
%   \parbox[c]{9em}{\includegraphics[width=9em]{./Figs/visconst/nn683fagaftrec2}}
%   \\
%  \parbox[c]{9em}{\includegraphics[width=9em]{./Figs/visconst/nn683fagaftpet2}}
%  \\
% \end{tabular} 
\end{tabular}
\end{tabular} 

%\vspace*{-0.001in}
\end{tabular}
  \end{adjustbox}
    \caption{\small{Visualization of a randomly selected neuron's inverted input attribution maps (IAMs) before fine-tuning and after fine-tuning in meta-testing. The model is obtained by {\RMAMLb}. The second row shows IAMs of the model before fine-tuning. The third row shows IAMs of the model after fine-tuning. One can find that the difference between the IAMs before fine-tuning and after fine-tuning is small, suggests that robust meta-update itself can provide the robustness adaptation without additional adversarial training.
}
  }
  \label{vis_fine_tune}
\end{figure}

Once obtain a model using {\RMAMLb}, we can test the impact of the standard fine-tuning on its robustness. Figure\,\ref{vis_fine_tune} shows a randomly selected neuron's inverted input attribution maps (IAMs) before standard fine-tuning and after standard fine-tuning in the meta-testing phase. The second row shows IAMs of the model before fine-tuning. The third row shows IAMs of the model after fine-tuning. One can find that the difference is small between the IAMs before fine-tuning and after fine-tuning, suggests that robust meta-update itself can provide the robustness adaptation without additional adversarial training.

% \subsection*{B: Details of Experiments}
% \label{experiments}

% \textbf{Dataset and Model} To test the effectiveness of our methods, we employ the dataset MiniImageNet \cite{vinyals2016matching}, which is the benchmark for few-shot learning. MiniImageNet contains 100 classes with 600 samples in each class. We use the training set with 64 classes and test set with 20 classes. In our experiments, we downsize each image to $84\times 84\times 3$. We choose a four-convolutional-layer CNN model for classification.

% \textbf{Setting} We consider the 1-shot (5-shot) 5-way image classification task, i.e., the inner-gradient update is implemented using five classes and one (five) fine-tuning images for each class in one task. Each batch contains four tasks. We set the number of gradient update steps $K = 5$. For the meta-update, we use 15 validation images for each class. We set the gradient step size in the fine-tuning as $\alpha=0.01$, and the step sizes in the meta-update as $\beta_1=0.001, \beta_2=0.001$ for clean images and adversarial images, respectively. The number of gradient update steps in meta-testing is set to $10$.

\section{Details of Experiments}\label{ssec4}

To test the effectiveness of our methods, we employ the MiniImageNet dataset \cite{vinyals2016matching}, which is the benchmark for few-shot learning. MiniImageNet contains 100 classes with 600 samples in each class. We use the training set with 64 classes and test set with 20 classes. In our experiments, we downsize each image to $84\times 84\times 3$.

we consider the 1-shot 5-way image classification task, i.e., the inner-gradient update (fine-tuning) is implemented using five classes and one fine-tuning image for each class in one single task. In meta-training, Each batch contains four tasks. We set the number of gradient update steps $K = 5$ in meta-training. For the meta-update, we use $15$ validation images for each class. We set the gradient step size in the fine-tuning as $\alpha=0.01$, and the gradient step sizes in the meta-update as $\beta_1=0.001, \beta_2=0.001$ for clean validation data and adversarial validation data, respectively.

\begin{figure}[htb]
\centerline{
\includegraphics[width=.55\textwidth,height=!]{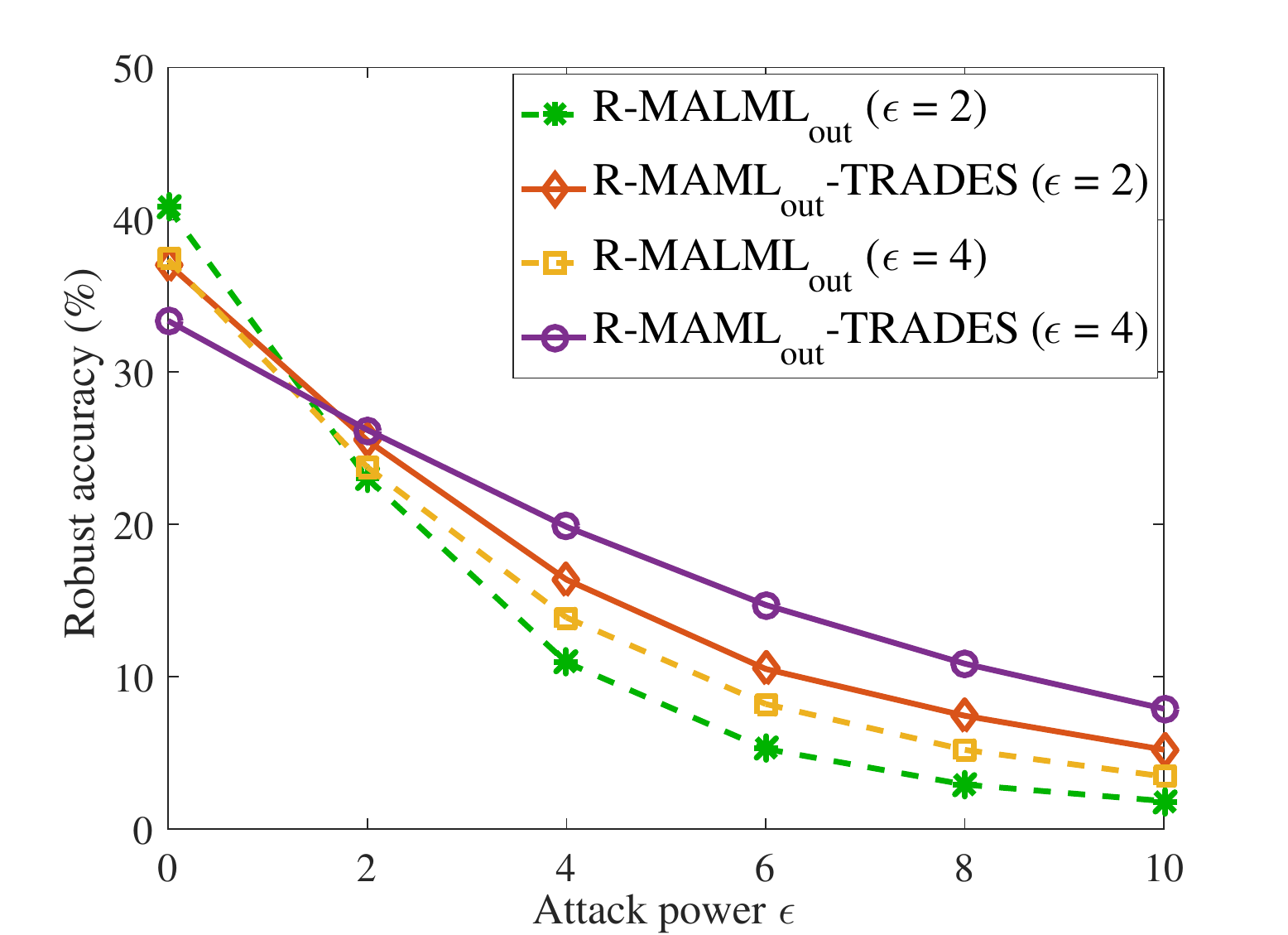}
}
\vspace*{-0.13in}
\caption{\small{
RA versus (testing-phase) PGD attacks at different values of perturbation strength $\epsilon$. Here the robust  models are trained by {\RMAMLbTr} and {\RMAMLb}. Each method trains two models under the training attack strength of $\epsilon=2, 4$, respectively.
Results show that {\RMAMLbTr} has the ability to defend stronger attacks than {\RMAMLb}.
}}
  \label{comp_dif_eps}
  %\vspace*{-0.2in}
\end{figure}

\section{Additional Comparisons on MiniImageNet}\label{ssec5}

Figure\,\ref{comp_dif_eps} shows robust accuracy (RA) performance of models trained using our methods. One can see that {\RMAMLbTr} has the ability to defend stronger attacks than {\RMAMLb}.

% \begin{figure}[htb]
% \centerline{
% \includegraphics[width=.55\textwidth,height=!]{./Figs/AdvAccComp}
% }
% \vspace*{-0.13in}
% \caption{\small{
% RA versus (testing-phase) PGD attacks at different values of perturbation strength $\epsilon$. Here the robust  models are trained by {\RMAMLbTr} and {\RMAMLb}. Each method trains two models under the training attack strength of $\epsilon=2, 4$, respectively.
% Results show that {\RMAMLbTr} has the ability to defend stronger attacks than {\RMAMLb}.
% }}
%   \label{comp_dif_eps}
%   %\vspace*{-0.2in}
% \end{figure}

In Table\,\ref{1shot_resnet}, we compare the SA/RA performance of variants of {\RMAMLb} including {\RMAMLb}(AT), the TRADES regularization with unlabeled data {\RMAMLbTr}, the version with contrastive learning {\RMAMLbCL}. One can see that {\RMAMLbCL} yields the best SA and RA among all meta-learning methods.

\begin{table}[h]

\caption{
SA/RA performance of different variants of proposed {\RMAMLb} under the 1-shot 5-way scenario on ResNet18.}
\begin{center}
\small
\label{1shot_resnet}
\resizebox{0.7\textwidth}{!}{
\begin{tabular}{l||c|c|c}
\hline
\hline
 & SA & RA \\
 \hline
MAML & 43.1\% & 5.347\%  \\
\hline
AQ \citep{goldblum2019adversarially} & 30.04\% & 20.05\%  \\
\hline
{\RMAMLb}(AT) (ours) & 38.94\% & 19.94\%   \\
% \hline
% {\RMAMLb}-CL & 41.73\% & 22.65\%  \\
% \hline
% {\RMAMLb}-CL (pretrained) & 42.67\% & 23.20\%  \\
\hline
{\RMAMLbTr} (ours) & 41.94\% & 20.19\%  \\
\hline
{\RMAMLbCL} (ours) & \bf{43.98\%} & \bf{21.47\%}  \\
\hline
\hline
\end{tabular}}
\end{center}
\end{table}

% \begin{table}[h]

% \caption{
% SA/RA performance of different variants of proposed {\RMAMLb} and baselines under the 5-shot 5-way scenario.}
% \begin{center}
% \small
% \label{5shot_ourmethods}
% \resizebox{0.7\textwidth}{!}{
% \begin{tabular}{l||c|c|c}
% \hline
% \hline
%  & SA & RA \\
%  \hline
% MAML & \bf{61.2\%} & 5.466\% \\
% \hline
% AQ \citep{goldblum2019adversarially} & 47.27\% & 38.02\% \\
% \hline
% {\RMAMLb} (ours) & 55.7\% & 37.01\%  \\
% \hline
% {\RMAMLb}(TRADES) (ours) & 54.44\% & 38.5\%  \\
% \hline
% {\RMAMLbTr} (ours) & 55.9\% & 39.14\%  \\
% % \hline
% % {\RMAMLbTr}-CL (ours) & 54.98\% & \bf{39.96\%}  \\
% \hline
% \hline
% \end{tabular}}
% \end{center}
% \end{table}

\section{Experiments on CIFAR-FS}\label{ssec6}

We also test our proposed methods on CIFAR-FS \citep{bertinetto2018meta}, which is an image classification dataset containing 64 classes of training data and 20 classes of evaluation data. The compared methods are the same as in Table\,\ref{tab: summary_CL}. We keep the settings to be the same as in the test on MiniImagenet except we set $\epsilon = 8$. To perform data augmentation in experiments, we mine $500$ additional unlabeled data for each training class from the STL-10 dataset \citep{coates2011analysis}.

Table\,\ref{cifar_1shot5way} and Table\,\ref{cifar_5shot5way} show the comparisons in 1-Shot 5-Way and 5-Shot 5-Way learning scenarios, respectively. One can see that our methods outperform the baseline methods MAML and AQ \citep{goldblum2019adversarially}. The results also indicate that semi-supervised learning (in terms of TRADES and contrastive learning) can further boost the performance. In particular, as shown by Table\,\ref{cifar_1shot5way} and Table\,\ref{cifar_5shot5way}, {\RMAMLbCL} leads to $10\%$ SA improvement and $3\%$ RA improvement compared to AQ under the MAML 1-Shot 5-Way setting, and $5.6\%$ SA improvement and $1.3\%$ RA improvement under the 5-Shot 5-Way setting.

\begin{table}[h]
\caption{
SA/RA performance of our proposed methods on CIFAR-FS \citep{bertinetto2018meta} (1-Shot 5-Way).
}

\begin{center}
\small
\label{cifar_1shot5way}
\resizebox{0.7\textwidth}{!}{
\begin{tabular}{l||c|c}
\hline
\hline
 & SA & RA \\
\hline
MAML & \bf{51.07\%} & 0.235\%    \\
\hline
AQ \citep{goldblum2019adversarially} & 31.25\% & 26.34\%  \\
\hline
{\RMAMLb}(AT) (ours) & 39.76\% & 26.15\%   \\
\hline
{\RMAMLb}(TRADES) (ours) & 40.23\% & 27.45\%   \\
\hline
{\RMAMLb}-TRADES (ours) & 40.59\% & 28.06\%   \\
\hline
{\RMAMLbCL} (ours) & 41.25\% & \bf{29.33\%} \\
\hline
\hline
\end{tabular}}
\end{center}
\end{table}

\begin{table}[h]
\caption{
SA/RA performance of our proposed methods on CIFAR-FS \citep{bertinetto2018meta} (5-Shot 5-Way).
}

\begin{center}
\small
\label{cifar_5shot5way}
\resizebox{0.7\textwidth}{!}{
\begin{tabular}{l||c|c}
\hline
\hline
 & SA & RA \\
\hline
MAML & \bf{67.2\%} & 0.225\%    \\
\hline
AQ \citep{goldblum2019adversarially} & 52.32\% & 33.96\%  \\
\hline
{\RMAMLb}(AT) (ours) &  57.18\% & 32.62\%   \\
\hline
{\RMAMLb}(TRADES) (ours) & 57.46\% & 34.72\%  \\
\hline
{\RMAMLb}-TRADES (ours) & 57.62\% & 34.76\% \\
\hline
{\RMAMLbCL} (ours) & 57.95\% & \bf{35.30\%} \\
\hline
\hline
\end{tabular}}
\end{center}
\end{table}

\section{Experiments on Omniglot}\label{ssec7}

\begin{figure}[h]
	\begin{minipage}[t]{0.5\linewidth}
		\centering
		\includegraphics[trim=0 0 0 0,clip,width=0.995\linewidth]{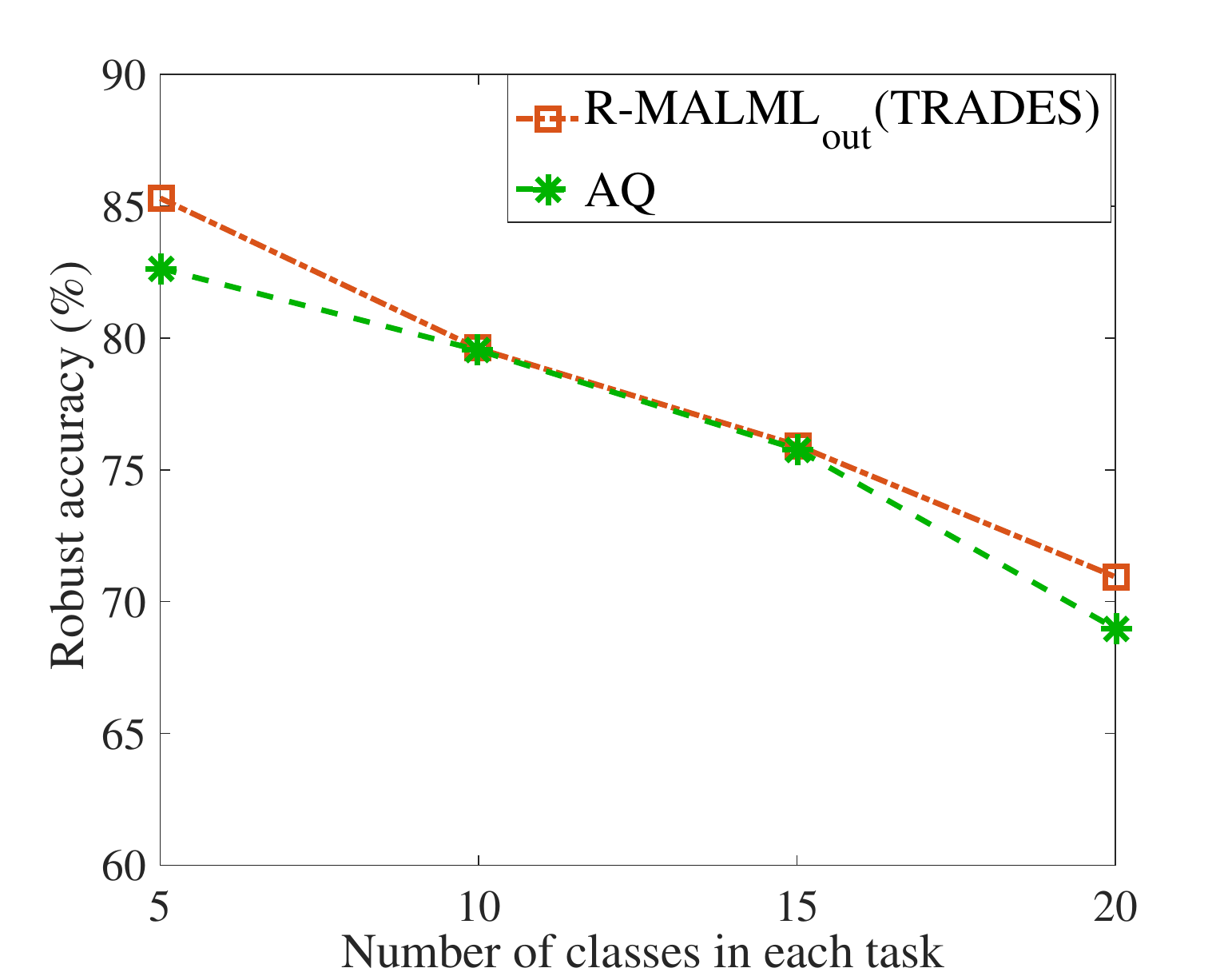}
					%\vspace{-0.2in}
					{\scriptsize \center (a) RA}
	\end{minipage}%
	\begin{minipage}[t]{0.5\linewidth}
		\centering
		\includegraphics[trim=0 0.2 0 0,clip,width=1\linewidth]{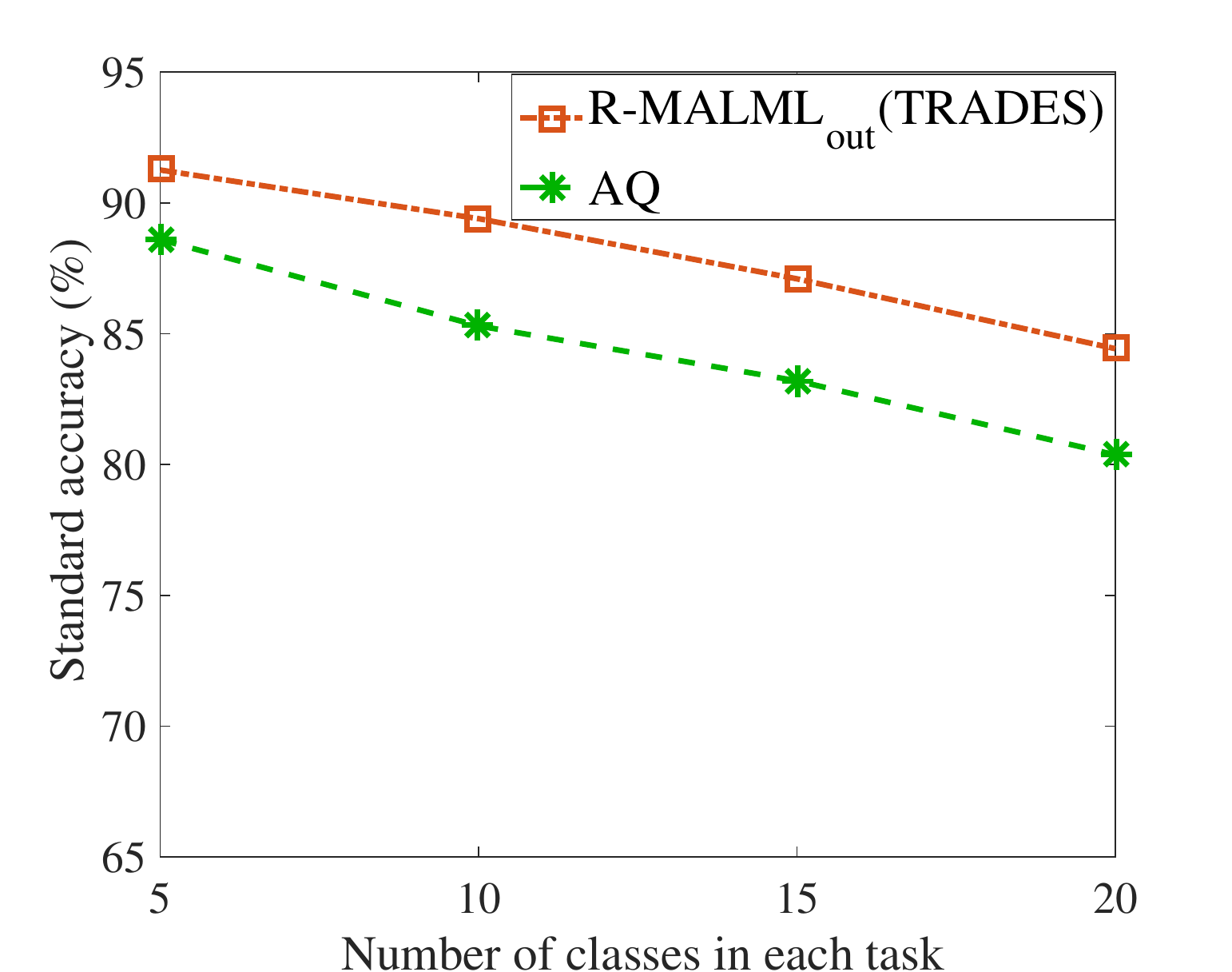}
					%\vspace{-0.2in}
					{\scriptsize \center (b) SA}
	\end{minipage}
	\caption{Performance of {\RMAMLb}(TRADES) and AQ \citep{goldblum2019adversarially} on Omniglot versus number of classes in each task (from $5$ to $20$ ways): (a) RA. (b) SA. }
	\label{omniglot_comp}
\end{figure}

We then conduct experiments on Omniglot \citep{lake2015human}, which includes handwritten characters from 50 different alphabets. There are 1028 classes of training data and 423 classes of evaluation data. Due to the hardness of finding the unlabeled data with similar patterns, we only test our supervised learning methods on Omniglot. %We compare {\RMAMLb}(AT) and {\RMAMLb}(TRADES) to MAML and AQ \citep{goldblum2019adversarially}. 
We compare {\RMAMLb}(TRADES) to AQ \citep{goldblum2019adversarially} in the 1-shot (5, 10, 15, 20)-Way settings. Figure.\,\ref{omniglot_comp} shows the results of RA/SA under $\epsilon = 10$. The results show that {\RMAMLb}(TRADES) can obtain better performance than AQ. %Table\,\ref{Omniglot_1shot5way} and Table\,\ref{omniglot_1shot20way} show the comparisons in 1-Shot 5-Way and 1-Shot 20-Way learning scenarios, respectively. The results show that our methods can obtain better performance than the baseline methods.

\end{document}

%% file: math_commands.tex
\usepackage{amsmath,amsfonts,bm}

\def\eqref#1{equation~\ref{#1}}
\def\1{\bm{1}}

\DeclareMathAlphabet{\mathsfit}{\encodingdefault}{\sfdefault}{m}{sl}
\SetMathAlphabet{\mathsfit}{bold}{\encodingdefault}{\sfdefault}{bx}{n}

\DeclareMathOperator*{\argmin}{arg\,min}